\documentclass[10pt,journal,epsfig]{IEEEtran}

\usepackage[dvips]{graphicx}
\usepackage{graphicx}
\usepackage{amssymb}
\usepackage{cite}
\usepackage{subfigure}
\usepackage{amsmath}
\usepackage{algorithm}
\usepackage{algorithmic}
\usepackage{multirow}
\usepackage{xcolor}
\usepackage{etoolbox}
\usepackage{upgreek}
\usepackage{tikz}

\allowdisplaybreaks[4]

\makeatletter \patchcmd{\@makecaption}{\scshape}{}{}{}\makeatother

\IEEEoverridecommandlockouts

\newcommand\copyrighttext{%
  \footnotesize \textcopyright 2022 IEEE. Personal use of this material is permitted.
  Permission from IEEE must be obtained for all other uses, in any current or future
  media, including reprinting/republishing this material for advertising or promotional
  purposes, creating new collective works, for resale or redistribution to servers or
  lists, or reuse of any copyrighted component of this work in other works.
  }
\newcommand\copyrightnotice{%
\begin{tikzpicture}[remember picture,overlay]
\node[anchor=south,yshift=10pt] at (current page.south) {\fbox{\parbox{\dimexpr\textwidth-\fboxsep-\fboxrule\relax}{\copyrighttext}}};
\end{tikzpicture}%
}

\begin{document}


\title{Confederated Learning: Federated Learning with Decentralized Edge Servers}

\author{Bin Wang, Jun Fang, Hongbin Li, ~\IEEEmembership{Fellow,~IEEE}, Xiaojun Yuan, and Qing
Ling
\thanks{Bin Wang, Jun Fang and Xiaojun Yuan are with the National Key Laboratory
of Science and Technology on Communications, University of
Electronic Science and Technology of China, Chengdu 611731, China,
Email: JunFang@uestc.edu.cn; xjyuan@uestc.edu.cn}
\thanks{Hongbin Li is with the Department of Electrical and Computer Engineering,
Stevens Institute of Technology, Hoboken, NJ 07030, USA, E-mail:
Hongbin.Li@stevens.edu}
\thanks{Qing Ling is with the School of Computer Science and Engineering, Sun Yat-Sen
University, Guangzhou, Guangdong 510006, China, and also with Peng
Cheng Laboratory, Shenzhen, Guangdong 518066, China, Email:
lingqing556@mail.sysu.edu.cn}
\thanks{This work was supported in part by the National Science
Foundation of China under Grants 61871091.} }

\maketitle
\copyrightnotice







\begin{abstract}
Federated learning (FL) is an emerging machine learning paradigm
that allows to accomplish model training without aggregating data
at a central server. Most studies on FL consider a centralized
framework, in which a single server is endowed with a central
authority to coordinate a number of devices to perform model
training in an iterative manner. Due to stringent communication
and bandwidth constraints, such a centralized framework has
limited scalability as the number of devices grows. To address
this issue, in this paper, we propose a ConFederated Learning
(CFL) framework. The proposed CFL consists of multiple servers, in
which each server is connected with an individual set of devices
as in the conventional FL framework, and decentralized
collaboration is leveraged among servers to make full use of the
data dispersed throughout the network. We develop an alternating
direction method of multipliers (ADMM) algorithm for CFL. The
proposed algorithm employs a random scheduling policy which
randomly selects a subset of devices to access their respective
servers at each iteration, thus alleviating the need of uploading
a huge amount of information from devices to servers. Theoretical
analysis is presented to justify the proposed method. Numerical
results show that the proposed method can converge to a decent
solution significantly faster than gradient-based FL algorithms,
thus boasting a substantial advantage in terms of communication
efficiency.
\end{abstract}





\section{Introduction}
In recent years, the rapid development of machine learning has
gained much attention in both the academia and the industry. The
tremendous success of machine learning is inseparable from the
help of huge data sets. Most conventional machine learning
algorithms are implemented in a centralized manner, requiring the
training data to be collected and processed in a central node.
However, securely aggregating heterogeneous data dispersed over
various data sources or organizations is a non-trivial task.
Processing the huge amount of data in a centralized fashion also
poses significant challenges for the data server. The challenges
concurrently arise from a privacy-protecting perspective. In some
data-sensitive areas such as the health care and financial
services, the confidentiality of users' data is of great concern
and should be protected. In such cases, sending users' data to a
centralized node may not be allowed.


Federated learning (FL) \cite{KonecnyMcMahan16} is a new
paradigm that enables model training without gathering data at
a central server. Such a merit makes it amiable for
data-intensive and privacy-sensitive machine learning
applications. So far most studies
\cite{KonecnyMcMahan16,PathakWainwright20,ZhangHong21,
NiuWei21,ZhouLi21,LiSahu20,McMahanMoore17,LiuChen20,KarimireddyKale20,HaddadpourMahdavi19}
focus on a \emph{centralized} FL framework, in which there is a
central server and a number of spatially distributed devices
(users). The server is bidirectionally connected to each user
which holds the data. To accomplish model training, FL employs a
computation-then-aggregation strategy. Specifically, in each
iteration, the central server first distributes the global model
to each user. Based on the global model, each user updates its
local model using its local data. The updated local model is then
uploaded to the server. At last, the server fuses the local models
to obtain a new global model. During this training process, the
data are preserved locally and only the training model is
exchanged, thus circumventing the need of gathering the data from
users to the central server.




Nevertheless, FL still faces challenges from both theoretical and
practical aspects. One fundamental problem of the single-server FL
system is poor scalability. Note that FL may operate in a wireless edge
network where the communication resource is severely constrained.
Due to limited bandwidth, at each iteration only a small subset of
users can be selected to interact with the server, which leads to
a low efficiency and also calls for a judiciously designed
scheduling policy \cite{YangLiu19,ShiZhou20, YangJiang20}. A line
of research to address the scalability issue is decentralized FL
\cite{HegedusDanner19,
LalithaKilinc19,XingSimeone20,SavazziNicoli20,NguyenDing20,WangJoshi21,JiangBalu17,
HaddadpourKamani19,KoloskovaStich19,WarnatSchultze21}, which has
attracted much interest due to their enhanced scalability as well
as its strengthened robustness to server failures. Typically,
decentralized FL is implemented on a decentralized network
consisting of a number of nodes. The decentralized network does
not have a global coordinator; instead, all nodes are connected in
a peer-to-peer manner. In these works, the nodes are assumed to be
the data-holders and thus the decentralized network forms a D2D
(Device-to-Device) network. Nevertheless, such a fully decentralized
setting may not fit in well with the current wireless edge network.

Another major challenge of FL is excessively high communication
overhead caused by frequent information exchange between the
server and the users. In many practical scenarios, communication
is much more costly than computation. It is, therefore, of vital
importance to reduce the communication overhead for FL. Many
existing studies \cite{McMahanMoore17,
LiSahu20,LiuChen20,LiuChen22,KarimireddyKale20,LiangShen19,YuanMa20,
HaddadpourMahdavi19,KoloskovaStich19} employ gradient descent or
proximal type of methods to perform training. These methods
require a massive amount of information exchanges because gradient
descent (with decreasing stepsizes) requires a large number of
iterations to converge. To relieve this issue, some works
\cite{McMahanMoore17,LiuChen20,LiuChen22,
KarimireddyKale20,LiangShen19,YuanMa20,HaddadpourMahdavi19}
suggest to run multiple iterations of local gradient descent
between adjacent aggregation steps. However, recent studies
\cite{LiHuang19} find that setting the number of local iterations
too large may have an unfavorable impact on the convergence speed.
Recently, more advanced optimization algorithms
\cite{ZhangHong21,PathakWainwright20,NiuWei21,PhamNguyen21} are
employed in FL. These works are mainly based on the ADMM
(alternating direction method of multipliers) algorithm, which
decomposes the original problem into a number of subproblems. In
general, ADMM type of algorithms require only a small number of
iterations to converge, thus having the potential to substantially
reduce the communication cost. Nevertheless, none of these
algorithms can be nontrivially extended to the CFL framework
considered in this work.

In this paper, we introduce a multi-server based FL framework,
whereby the servers form a decentralized network while each server
is connected to an individual set of edge devices. Such a
framework is a union of sovereign servers united for the purpose
of learning a global model, and thus is referred to as
confederated learning (CFL). CFL can better address the
scalability issue than the centralized one. Meanwhile, it does not
involve complex network management required by the D2D network.
Note that it is reasonable to assume the servers to work in a
decentralized manner since there may not be a global center to
coordinate these servers. In addition, the intelligent nature of
B5G and 6G networks calls for extensive and flexible
self-organizations of local or trans-regional cooperations. We
note that confederated learning was introduced in \cite{LiuFox19}
as a term to characterize FL with ``vertically separated'' data,
e.g., different data types (lab tests, diagnosis, medications,
treatments, etc.) of a given patient are located at different
locations and cannot be easily matched with each other. Although
using the same term, the meaning of CFL in this work is totally
different from that of \cite{LiuFox19}.

Within this framework, we develop an efficient ADMM-based CFL
algorithm. The proposed ADMM algorithm is characterized with two
distinctive features. Firstly, to alleviate the need of uploading
a huge amount of information from massive distributed devices to
each server, a random scheduling policy is employed, whereby each
device, at each iteration, is randomly activated with a small
probability and participates in the training process. Secondly,
considering the fact that subproblems of ADMM may not have a
closed-form solution, the proposed ADMM allows the subproblem to
be solved up to a certain accuracy. Theoretical analysis reveals
that the proposed algorithm enjoys a sublinear convergence rate.
Numerical results show that the proposed method can converge to a
decent solution significantly faster (i.e. with much fewer
communication rounds) than those gradient-based CFL algorithms,
thus presenting a substantial advantage in terms of communication
efficiency.

The rest of this paper is organized as follows. Some preliminaries
on convex functions are first introduced in Section
\ref{sec-preliminaries}. Then in Section \ref{sec-HFL}, we present
a CFL framework and formulate the CFL problem. A new ADMM
algorithm is proposed in Section \ref{sec-HADMM}. The convergence
result of the proposed algorithm and its proof are provided in
Section \ref{sec-convergence} and Section \ref{sec-proof},
respectively. Simulations results are provided in Section
\ref{sec-simulation}, followed by concluding remarks in Section
\ref{sec-conclusion}.

\section{Preliminaries}
\label{sec-preliminaries}
\subsection{Properties of Convex Functions} The subgradient of a
convex function $f$ is denoted as $\partial f$. If $f$ is
continuously differentiable, then we have $\partial f=\nabla f$.
For a convex function $f$, it always holds that
\begin{align}
&f(\boldsymbol{x})\geq f(\boldsymbol{y})+\langle \partial
f(\boldsymbol{y}), \boldsymbol{x}-\boldsymbol{y}\rangle, \forall
\boldsymbol{x},\boldsymbol{y},
\nonumber\\
& f(\textstyle\sum\limits_{t=1}^T \delta_t\boldsymbol{x}_t)
\leq\textstyle \sum\limits_{t=1}^T \delta_t f(\boldsymbol{x}_t), \
\text{if} \ \sum\limits_{t=1}^T\delta_t=1 \ \text{and} \
\delta_t\geq 0, \forall \boldsymbol{x}_t. \label{sec2-2-1}
\end{align}
where the second inequality is known as the Jensen's inequality. A
function $f$ is said to be $\mu$-strongly convex if it satisfies
\begin{align}
\textstyle f(\boldsymbol{x})\geq f(\boldsymbol{y})+\langle\partial
f(\boldsymbol{y}),\boldsymbol{x}-\boldsymbol{y}\rangle+\frac{\mu}{2}
\|\boldsymbol{x}-\boldsymbol{y}\|_2^2, \ \forall
\boldsymbol{x},\boldsymbol{y}. \label{def-strong}
\end{align}


\subsection{Commonly Used Inequalities} Given a triple of
arbitrary vectors $\boldsymbol{x}$, $\boldsymbol{y}$ and
$\boldsymbol{z}$, it holds
\begin{align}
2\langle\boldsymbol{x}-\boldsymbol{y},\boldsymbol{x}-\boldsymbol{z}\rangle
=\|\boldsymbol{x}-\boldsymbol{y}\|_2^2+\|\boldsymbol{x}-\boldsymbol{z}\|_2^2
-\|\boldsymbol{y}-\boldsymbol{z}\|_2^2 \label{sec2-2-2}
\end{align}
Meanwhile, for $\forall \boldsymbol{x},\boldsymbol{y}$, it holds
\begin{align}
2\langle\boldsymbol{x},\boldsymbol{y}\rangle\leq
\omega\|\boldsymbol{x}\|_2^2 +\omega^{-1}\|\boldsymbol{y}\|_2^2, \
\forall \omega>0. \label{sec2-2-2-1}
\end{align}

\begin{figure*}[!t]
    \centering
    \subfigure[]{\includegraphics[width=2.4in]{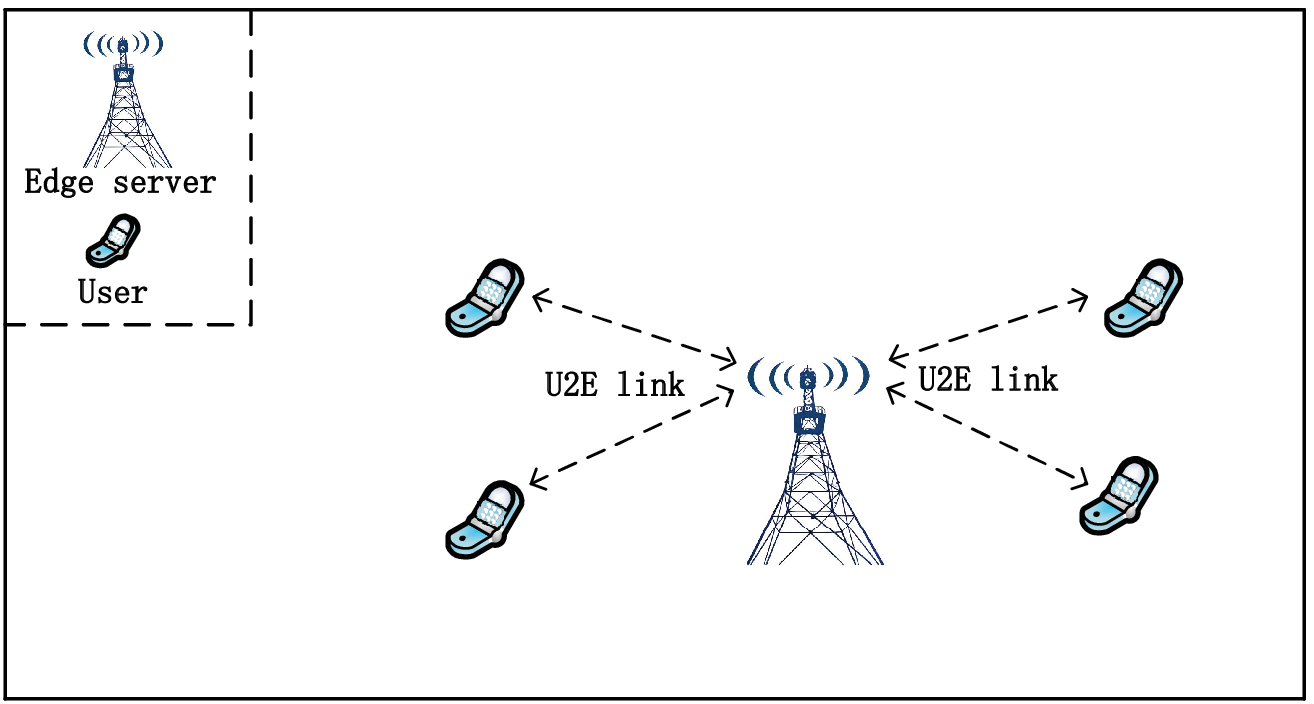}} \hfil
    \subfigure[]{\includegraphics[width=2.4in]{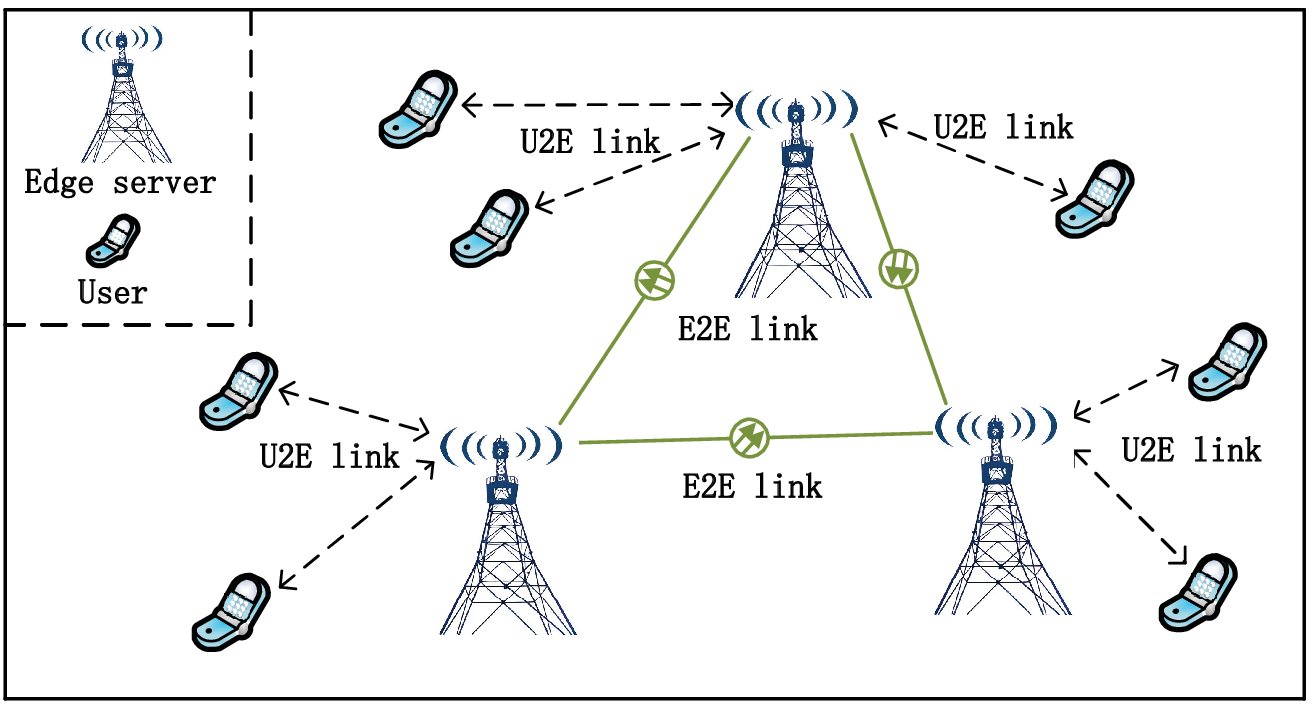}} \hfil
    \caption{(a): Conventional FL framework with a single ES;
    (b): Proposed CFL framework with multiple ESs.}
    \label{fig11}
\end{figure*}

\section{Confederated Learning}
\label{sec-HFL}
\subsection{CFL Framework}
\label{sec-intro-hflsystem} We consider a CFL framework consisting
of $l$ edge servers (ESs), in which the $i$th ES is connected to
$|S_i|$ edge devices (i.e. users) which hold the data. Here $S_i$
represents the set of users served by the $i$th ES and $|S_i|$ is
the cardinality of $S_i$. Let $u_{ij}$ denote the $j$th user
served by the $i$th ES. It is assumed that the sets of users
served by different ESs are disjoint. Each ES can communicate with
its own users, while communications among users are not allowed.
Also, ESs form a decentralized network that can be abstracted as a
graph $G=\{V,E\}$, in which there is no global coordinator and
each ES is only allowed to communicate with its neighboring ESs.
Clearly, the conventional single ES-based FL framework is a
special case of the CFL framework (see Fig. \ref{fig1}). The CFL
framework also covers the centralized multi-ES system
\cite{WangWang20} as a special case, where the ESs form a
star-type communication network.

The CFL framework is different from the peer-to-peer FL
\cite{HegedusDanner19,LalithaKilinc19,XingSimeone20,
SavazziNicoli20}. The CFL system is more suitable for applications
residing on wireless edge networks while the peer-to-peer FL is
more suitable for D2D networks. A recent work
\cite{MaNikolakopoulos18} proposed an in-network acceleration
scheme by appointing a portion of nodes to be the (virtual) local
fusion centers. However, the communication pattern still follows a
fully decentralized manner.

\subsection{Problem Formulation}
Consider the following optimization problem:
\begin{align}
\textstyle\mathop {\min }\limits_{\boldsymbol{x}\in\mathbb{R}^d} \ \sum_{i=1}^l
f_i(\boldsymbol{x})=\sum_{i=1}^{l}\sum_{j=1}^{|S_i|}f_{ij}(\boldsymbol{x};
\mathcal{D}_{ij}),
\label{sec1-1}
\end{align}
where
$f_i(\boldsymbol{x})\triangleq\sum_{j=1}^{|S_i|}f_{ij}(\boldsymbol{x};
\mathcal{D}_{ij})$, $f_{ij}(\boldsymbol{x};\mathcal{D}_{ij})$ is a
convex, proper and lower semi-continuous function held by user
$u_{ij}$ and $\mathcal{D}_{ij}$ represents the local data set
stored at user $u_{ij}$. For a learning task, the variable
$\boldsymbol{x}\in\mathbb{R}^n$ represents the global model
parameter vector that is to be learned. The function $f_{ij}$ is
referred to as the local loss function. If $l=1$, then
(\ref{sec1-1}) degenerates into the standard FL problem. By
introducing a set of auxiliary variables $\{\boldsymbol{y}_i\}$,
we can reformulate (\ref{sec1-1}) into the following problem:
\begin{align}
\mathop {\min }\limits_{\{\boldsymbol{x}_{ij}\}_{i,j},\{\boldsymbol{y}_{i}\}_i} & \
\textstyle\sum_{i=1}^{l}\sum_{j=1}^{|S_i|}f_{ij}(\boldsymbol{x}_{ij};\mathcal{D}_{ij})
\nonumber\\
\text{s.t.} & \ \boldsymbol{x}_{ij}=\boldsymbol{y}_{i}, \quad
\forall i\in\{1,\ldots,l\}, \forall j\in S_i,
\nonumber\\
& \ \boldsymbol{y}_{1}=\boldsymbol{y}_{2}=\cdots=\boldsymbol{y}_{l}.
\label{sec1-2}
\end{align}
where $\boldsymbol{x}_{ij}\in\mathbb{R}^{n}$ is the local variable
held by user $u_{ij}$ and $\boldsymbol{y}_i\in\mathbb{R}^{n}$ is
the local variable held by the $i$th ES. In (\ref{sec1-2}), the
first equality constraint, i.e.,
$\boldsymbol{x}_{ij}=\boldsymbol{y}_{i}$, forces the consistency
between the $i$th ES's local variable and those of its users. The
second constraint forces the local variables of the ESs to be
equal to each other. Clearly, (\ref{sec1-2}) is essentially the
same as (\ref{sec1-1}). Nevertheless, (\ref{sec1-2}) can not be
solved in a decentralized manner since tackling the second
constraint demands centralized operations. To circumvent this
obstacle, we resort to solving the following equivalent problem:
\begin{align}
&\text{CFL Optimization:}
\nonumber\\
&\qquad \quad\mathop {\min
}\limits_{\{\boldsymbol{x}_{ij}\}_{i,j}, \{\boldsymbol{y}_{i}\}_i}
\ \textstyle\sum_{i=1}^{l}\sum_{j=1}^{|S_i|}
f_{ij}(\boldsymbol{x}_{ij};\mathcal{D}_{ij})
\nonumber\\
&\qquad\qquad\qquad\qquad\qquad\text{s.t.}  \
\boldsymbol{x}_{ij}=\boldsymbol{y}_{i}, \forall
i\in\{1,\ldots,l\}, \forall j\in S_i,
\nonumber\\
&\qquad\qquad \qquad\qquad\quad\qquad\ \boldsymbol{A}\boldsymbol{y}=\boldsymbol{0}.
\label{sec1-4}
\end{align}
where $\boldsymbol{y}\in\mathbb{R}^{ln}$ denotes the vertical
stack of $\boldsymbol{y}_i$s, i.e.,
$\boldsymbol{y}\triangleq[\boldsymbol{y}_1;
\boldsymbol{y}_2;\cdots;\boldsymbol{y}_l]$,
$\boldsymbol{A}\triangleq \boldsymbol{A}_{in} \otimes
\boldsymbol{I}_n$, $\boldsymbol{A}_{in}$ is the incidence matrix
of the graph $G$, $\otimes$ denotes Kronecker product and
$\boldsymbol{I}_n$ is an $n\times n$ identity matrix. It is
well-known that
$\boldsymbol{A}\boldsymbol{y}=\boldsymbol{0}\Leftrightarrow
\boldsymbol{y}_{1}=\boldsymbol{y}_{2}=\cdots=\boldsymbol{y}_{l}$.
Thus (\ref{sec1-4}) is also equivalent to (\ref{sec1-1}). Notably,
previous research \cite{WangFang18,ShiLing15b,ShiLing14} on
decentralized optimization has paved a way on how to handle the
second constraint in a decentralized manner. To ease subsequent
expositions, hereafter we omit $\mathcal{D}_{ij}$ in $f_{ij}$.



\subsection{Communication Bottleneck and Random Scheduling}
In our proposed CFL framework, there exists two types of data
transmissions, namely, user-to-ES (U2E) communications and
ES-to-ES (E2E) communications. Generally, for CFL, the
communication bottleneck lies in the U2E communications. This is
because each ES may be assigned with a large number of users. Thus
sending the local update from each user to its associated ES
consumes a significant amount of communication resource and
meanwhile may incur a high latency. To overcome this difficulty,
in our algorithm, we randomly choose a small subset of users at
each iteration to communicate with its ES. Specifically, each user
is assigned a same probability $\alpha$, and is independently
activated with probability $\alpha$ at each iteration to report
its local update to its associated ES. This user selection policy
is termed as a random scheduling policy. Such a policy allows each
user to have the same chance to access its associated ES.
Meanwhile, at each iteration only a small number of users are
activated to access ESs, which enables the algorithm to operate
under stringent communication and delay constraints.

\section{Proposed Algorithm}
\label{sec-HADMM} In this section, we propose a new ADMM algorithm
that can accommodate the CFL framework. Some discussions are then
provided to shed some insight into the proposed algorithm.

\subsection{Algorithm Development}
To facilitate subsequent expositions, we first introduce the
following notations:
\begin{align}
&\boldsymbol{x}\triangleq[\boldsymbol{x}_1;\cdots;\boldsymbol{x}_l],
\boldsymbol{x}_i\triangleq[\boldsymbol{x}_{i1};\cdots;\boldsymbol{x}_{i|S_i|}],
\boldsymbol{\lambda}\triangleq[\boldsymbol{\lambda}_1;\cdots;\boldsymbol{\lambda}_l],
\nonumber\\
&\boldsymbol{\lambda}_i\triangleq[\boldsymbol{\lambda}_{i1};\cdots;
\boldsymbol{\lambda}_{i|S_i|}],\partial f(\boldsymbol{x})
\triangleq[\partial f_1(\boldsymbol{x}_1);\cdots;\partial
f_l(\boldsymbol{x}_l)],
\nonumber\\
&\partial f_i(\boldsymbol{x}_i)\triangleq[\partial
f_{i1}(\boldsymbol{x}_{i1}); \cdots;\partial
f_{i|S_i|}(\boldsymbol{x}_{i|S_i|})],\bar{\boldsymbol{\lambda}}
\triangleq[\bar{\boldsymbol{\lambda}}_1;\cdots;\bar{\boldsymbol{\lambda}}_l],
\nonumber\\
&\boldsymbol{y}\triangleq[\boldsymbol{y}_1;\cdots;\boldsymbol{y}_l],
\boldsymbol{H}=\text{bldig}\{\boldsymbol{H}_1;\cdots;\boldsymbol{H}_l\},
\label{sec3-2-1}
\end{align}
where $\boldsymbol{H}_i\in\mathbb{R}^{n\times n|S_i|}$ is a matrix
obtained by concatenating $|S_i|$ identity matrices of size
$n\times n$.

The augmented Lagrangian function of (\ref{sec1-4}) is given as
\begin{align}
&L_A\big(\big\{\{\boldsymbol{x}_{ij},\boldsymbol{\lambda}_{ij}\}_{j=1}^{|S_i|},
\boldsymbol{y}_{i}\big\}_{i=1}^l,\boldsymbol{\beta}\big)
\nonumber\\
=&\textstyle\sum_{i=1}^{l}\sum_{j=1}^{|S_i|}
\big( f_{ij}(\boldsymbol{x}_{ij})+\langle\boldsymbol{\lambda}_{ij},\boldsymbol{x}_{ij}-
\boldsymbol{y}_{i}\rangle+\frac{\sigma_1}{2}\|\boldsymbol{x}_{ij}-
\boldsymbol{y}_{i}\|_2^2\big)
\nonumber\\
&\textstyle+\langle\boldsymbol{\beta},\boldsymbol{A}\boldsymbol{y}\rangle+
\frac{\sigma_2}{2}\|\boldsymbol{A}\boldsymbol{y}\|_2^2
\end{align}
where $\{\boldsymbol{\lambda}_{ij}\}$ and $\boldsymbol{\beta}$ are
Lagrangian multipliers, $\sigma_1$ and $\sigma_2$ are man-crafted
parameters. Based on $L_A$, we can easily deduce a standard ADMM
algorithm as follows:
\begin{align}
&\textstyle\boldsymbol{x}_{ij}^{k+1}=\arg\min\limits_{\boldsymbol{x}_{ij}}
\
f_{ij}(\boldsymbol{x}_{ij})+\frac{\sigma_1}{2}\|\boldsymbol{x}_{ij}-
\boldsymbol{y}_{i}^k+\frac{\boldsymbol{\lambda}_{ij}^k}{\sigma_1}\|_2^2,
\nonumber\\
&\boldsymbol{y}^{k+1}=\arg\min\limits_{\boldsymbol{y}} \
\textstyle\sum\nolimits_{i=1}^{l}\sum\nolimits_{j=1}^{|S_i|}
\big( \frac{\sigma_1}{2}\|\boldsymbol{x}_{ij}^{k+1}-\boldsymbol{y}_{i}+
\sigma_1^{-1}\boldsymbol{\lambda}_{ij}^k\|_2^2\big)
\nonumber\\
&\qquad\qquad\qquad\quad\textstyle+\frac{\sigma_2}{2}\|\boldsymbol{A}
\boldsymbol{y}+\sigma_2^{-1}\boldsymbol{\beta}^{k}\|_2^2,
\nonumber\\
&\boldsymbol{\lambda}_{ij}^{k+1}=\boldsymbol{\lambda}_{ij}^{k}+
\sigma_1(\boldsymbol{x}_{ij}^{k+1}-\boldsymbol{y}_{i}^{k+1}),
\ \forall i, \forall j\in S_i,
\nonumber\\
&\boldsymbol{\beta}^{k+1}=\boldsymbol{\beta}^{k+1}+
\sigma_2\boldsymbol{A}\boldsymbol{y}^{k+1}.
\label{sec2-1}
\end{align}
Nevertheless, the above algorithm can not fulfil our needs since,
firstly, this algorithm requires all users to participate in the
$\boldsymbol{x}_{ij}^{k+1}$-update and send their local updates to
their respective ESs, which incurs a prohibitively high
communication cost. Secondly, the algorithm demands an exact
solution of the $\boldsymbol{x}_{ij}^{k+1}$-subproblem. This is a
stringent requirement since obtaining the exact solution of an
optimization problem might be computationally expensive. Thirdly,
the $\boldsymbol{y}^{k+1}$-subproblem can not be solved in a
decentralized manner since $\|\boldsymbol{A}\boldsymbol{y}\|_2^2$
is a nonseparable term.

To address the above difficulties, we propose a new ADMM
algorithm, which is summarized in Algorithm \ref{alg:1}.
Specifically, in each iteration of Algorithm \ref{alg:1}, only a
subset of users are selected (with probability $\alpha$) to
participate in the $\boldsymbol{x}^{k+1}$-update, thus avoiding the
need of data transmissions from every user to its ES. Meanwhile,
Algorithm \ref{alg:1} allows the
$\boldsymbol{x}_{ij}^{k+1}$-subproblem to be solved up to an
$\epsilon$-accuracy instead of solving it exactly. Lastly, in the
proposed algorithm, we use a judiciously designed extra proximal
term such that the $\boldsymbol{y}^{k+1}$-subproblem can be solved
in a decentralized manner.

With the notations defined in (\ref{sec3-2-1}), the update of
$\bar{\boldsymbol{\lambda}}_{ij}^{k+1}$ and
$\boldsymbol{\lambda}_{ij}^{k+1}$ in Algorithm \ref{alg:1} can be
compactly written as
\begin{align}
&\bar{\boldsymbol{\lambda}}^{k+1}=
\boldsymbol{\lambda}^k+\sigma_1(\boldsymbol{x}^{k+1}-\boldsymbol{H}^T
\boldsymbol{y}^{k+1}),
\nonumber\\
&\boldsymbol{\lambda}^{k+1}=\boldsymbol{\lambda}^k+\alpha
(\bar{\boldsymbol{\lambda}}^{k+1}-\boldsymbol{\lambda}^{k}).
\label{sec3-2-1-1}
\end{align}
The $\boldsymbol{y}^{k+1}$-subproblem, $k+1<\bar{k}$, can also be
compactly written as
\begin{align}
\boldsymbol{y}^{k+1}=\arg\min\limits_{\boldsymbol{y}} \
&\textstyle\frac{\alpha
\sigma_1}{2}\|\boldsymbol{x}^{k+1}-\boldsymbol{H}^T\boldsymbol{y}+
\frac{\boldsymbol{\lambda}^k}{\alpha\sigma_1}\|_2^2+
\nonumber\\
&\textstyle\frac{\sigma_2}{2}\|\boldsymbol{A}\boldsymbol{y}+
\frac{\boldsymbol{\beta}^{k}}{\sigma_2}\|_2^2
+\frac{\sigma_2}{2}\|\boldsymbol{y}-\boldsymbol{y}^k\|_{\alpha^{-1}
\boldsymbol{P}}^2. \label{sec3-2-1-2}
\end{align}

\begin{algorithm}
\caption{CFL-ADMM} \label{alg:1}
\begin{algorithmic}
\STATE{\textbf{Inputs}: parameters $\sigma_1$ and $\sigma_2$,
the activation probability $\alpha$ and the maximum number of
iterations $\bar{k}$. All initial vectors are set to
$\boldsymbol{0}$}. \STATE{\textbf{While} $(k+1)\leq \bar{k}$
\textbf{do}} \STATE{\textcircled{1} \ \textbf{User selection}:
Each user has a probability of $\alpha$ to be selected. The index
set of the users selected by ES $i$ in the $(k+1)$th iteration is
denoted as $I_i^{k+1}$.} \STATE{\textcircled{2} \ \textbf{Users
solve}:}
\begin{align}
&\left\{\begin{array}{ll}
\boldsymbol{x}_{ij}^{k+1}\overset{\epsilon^{k+1}}{\approx}\arg\min\limits_{\boldsymbol{x}_{ij}}
\ f_{ij}(\boldsymbol{x}_{ij})+\frac{\sigma_1}{2}\|\boldsymbol{x}_{ij}-\boldsymbol{y}_{i}^k
+\frac{\boldsymbol{\lambda}_{ij}^k}{\sigma_1}\|_2^2,
\\
\qquad\qquad\qquad\qquad \forall i, \forall j\in I_i^{k+1},
\\
\boldsymbol{x}_{ij}^{k+1}=\boldsymbol{x}_{ij}^{k}, \ \forall i, \forall j\notin
I_i^{k+1},
\end{array} \right.
\label{sec2-2-x}
\end{align}
\STATE{\textcircled{3} \ \textbf{Model upload}: Selected users
upload their local variables to ES $i$.} \STATE{\textcircled{4} \
\textbf{ESs solve}:}
\begin{align}
&\left\{ \begin{array}{ll}
\boldsymbol{y}^{k+1}=\arg\min\limits_{\boldsymbol{y}}\sum\limits_{i=1}^{l}
\sum\limits_{j=1}^{|S_i|}\frac{\alpha\sigma_1}{2}\|\boldsymbol{x}_{ij}^{k+1}-
\boldsymbol{y}_{i}+\frac{\boldsymbol{\lambda}_{ij}^k}{\alpha\sigma_1}\|_2^2+
\\
\quad\quad\frac{\sigma_2}{2}\|\boldsymbol{A}\boldsymbol{y}+\frac{\boldsymbol{\beta}^{k}}
{\sigma_2}\|_2^2+\frac{\sigma_2}{2}\|\boldsymbol{y}-\boldsymbol{y}^k
\|_{\alpha^{-1}\boldsymbol{P}}^2, \ k+1< \bar{k},
\\
\boldsymbol{y}^{\bar{k}}=\arg\min\limits_{\boldsymbol{y}}\sum\limits_{i=1}^{l}
\sum\limits_{j=1}^{|S_i|} \frac{\sigma_1}{2}\|\boldsymbol{x}_{ij}^{\bar{k}}
-\boldsymbol{y}_{i}+\frac{\boldsymbol{\lambda}_{ij}^{\bar{k}-1}}{\sigma_1}\|_2^2+
\\
\quad\quad\frac{\sigma_2}{2\alpha}\|\boldsymbol{A}\boldsymbol{y}+
\frac{\alpha\boldsymbol{\beta}^{\bar{k}-1}}{\sigma_2}\|_2^2+\frac{\sigma_2}{2}
\|\boldsymbol{y}-\boldsymbol{y}^{\bar{k}-1}\|_{\boldsymbol{P}}^2, \ k+1=\bar{k},
\\
\boldsymbol{\beta}^{k+1}=\boldsymbol{\beta}^{k}+\sigma_2\boldsymbol{A}
\boldsymbol{y}^{k+1},\ \text{if} \ (k+1)<\bar{k},
\\
\boldsymbol{\beta}^{\bar{k}}=\boldsymbol{\beta}^{\bar{k}-1}+
\frac{\sigma_2}{\alpha}\boldsymbol{A}\boldsymbol{y}^{\bar{k}}, \ \text{if} \
(k+1)=\bar{k}.
\end{array} \right.
\label{sec2-2-y}
\end{align}
\STATE{\textcircled{5} \ \textbf{Model download}: Each ES broadcasts its local
variable to its serving users.}
\STATE{\textcircled{6} \ \textbf{Users update}: }
\begin{align}
&\bar{\boldsymbol{\lambda}}_{ij}^{k+1}=\boldsymbol{\lambda}_{ij}^{k}+\sigma_1
(\boldsymbol{x}_{ij}^{k+1}-\boldsymbol{y}_{i}^{k+1}), \ \forall i, \forall j\in S_i,
\nonumber\\
&\boldsymbol{\lambda}_{ij}^{k+1}=\boldsymbol{\lambda}_{ij}^k+
\alpha(\bar{\boldsymbol{\lambda}}_{ij}^{k+1}
-\boldsymbol{\lambda}_{ij}^k), \ \forall i, \forall j\in S_i,
\label{sec2-2}
\end{align}
\par{\textbf{End While};}
\par{\textbf{Outputs: $\boldsymbol{x}_{ij}^{k+1}$};}
\end{algorithmic}
\end{algorithm}


\subsection{Training Process and Communication Efficiency}
\subsubsection{Training Process}
At each iteration of Algorithm \ref{alg:1}, the $i$th ES first
distributes $\boldsymbol{y}_i^{k}$ to its associated users. Then
the selected users update their local models by solving the
$\boldsymbol{x}_{ij}^{k+1}$-subproblem, $\forall j\in I_i^{k+1}$,
where $I_i^{k+1}$ denotes the index set of the users selected by
ES $i$ at the $(k+1)$th iteration. After the local update, the
selected users upload $\boldsymbol{x}_{ij}^{k+1}$ to its
associated ES. Then the ESs collaboratively solve the
$\boldsymbol{y}^{k+1}$-subproblem through local information
exchange. As will be shown later, the
$\boldsymbol{y}^{k+1}$-subproblem admits a closed-form solution.
Solving the $\boldsymbol{y}^{k+1}$-subproblem only needs to
exchange information among neighboring ESs once, which does not
incur additional latency and communication costs. It should be
noted that solving the $\boldsymbol{y}^{k+1}$-subproblem also
involves the local model parameters of those unselected users.
Nevertheless, since we have
$\boldsymbol{x}_{ij}^{k+1}=\boldsymbol{x}_{ij}^{k}$ for those
$j\notin I_i^{k+1}$, we can use the model parameters obtained in
the previous iteration for these unselected users. For this
purpose, each ES can build a history database to store its users'
model parameters obtained in the previous iteration. At last, the
update of $\boldsymbol{\lambda}_{ij}^{k+1}$ can be conducted
locally at each user.


\subsubsection{Communication Overhead Analysis}
\label{sec-overhead} At the $(k+1)$th iteration, each ES needs to
broadcast its local variable $\boldsymbol{y}_i^{k}$ to its users,
and each selected user uploads its local variable
$\boldsymbol{x}_{ij}^{k+1}$ to its associated ES. Since each user
is selected with a same probability $\alpha$, the average number
of users that participate the uplink U2E transmission at each
iteration is $\alpha\sum_{i=1}^{l}|S_i|$. As for the E2E
communication, each ES needs to communicate with its one-hop
neighboring ESs only once at each iteration. Overall, in an
average sense, the total number of messages that are exchanged
between ESs and between ESs and users is up to
$2l+\alpha\sum_{i=1}^{l}|S_i|$ at each iteration.



\subsection{Implementations and Discussions}
\subsubsection{Implementations of the $\boldsymbol{x}_{ij}^{k+1}$-subproblem}
In the $\boldsymbol{x}_{ij}^{k+1}$-subproblem, the notation
$\overset{\epsilon^{k+1}}{\approx}$ means that this problem is
solved up to an $\epsilon^{k+1}$-accuracy, i.e. the gradient of
the objective function satisfies
$\|\boldsymbol{\tau}_{ij}^{k+1}\|_2\leq \epsilon^{k+1}$, where
\begin{align}
\boldsymbol{\tau}_{ij}^{k+1}=\partial f_{ij}(\boldsymbol{x}_{ij}^{k+1})+
\boldsymbol{\lambda}_{ij}^k+\sigma_1(\boldsymbol{x}_{ij}^{k+1}-\boldsymbol{y}_{i}^k)
\label{imple-1}
\end{align}
Such a metric can be conveniently evaluated as the gradient
descent-based method is commonly used in solving the local
subproblem. Note that the $\epsilon$-accuracy is widely used in
existing literatures, e.g., \cite{ZhangHong21}. If $\epsilon$ is
set to $0$, then the subproblem should be solved exactly.


\subsubsection{Implementations of the $\boldsymbol{y}^{k+1}$-subproblem}
In the $\boldsymbol{y}^{k+1}$-subproblem, an extra proximal term
is added to enable the decentralized implementation. Here
$\boldsymbol{P}\in\mathbb{R}^{ln\times ln}$ is chosen to be
\begin{align}
\alpha^{-1}\boldsymbol{P}=\boldsymbol{D}-\boldsymbol{A}^T\boldsymbol{A},
\label{imple-2}
\end{align}
where $\boldsymbol{D}\in\mathbb{R}^{ln\times ln}$ is a diagonal
matrix whose choice will be elaborated in Section
\ref{sec-3}. It can be readily verified that the
$\boldsymbol{y}^{k+1}$-subproblem, $k+1<\bar{k}$, i.e.,
(\ref{sec3-2-1-2}), admits a closed-form solution given as
\begin{align}
\boldsymbol{y}^{k+1}=&\textstyle(\alpha\sigma_1\boldsymbol{H}\boldsymbol{H}^T
+\sigma_2\boldsymbol{D})^{-1}\Big(\alpha\sigma_1\boldsymbol{H}
(\boldsymbol{x}^{k+1}+\frac{\boldsymbol{\lambda}^k}
{\alpha\sigma_1})-
\nonumber\\
&\boldsymbol{A}^T\boldsymbol{\beta}^k+\sigma_2\big(\boldsymbol{D}-
\boldsymbol{A}^T\boldsymbol{A}\big)\boldsymbol{y}^k\Big)
\label{imple-2-1}
\end{align}
Note that the term $\boldsymbol{A}^T\boldsymbol{\beta}^k$ in
(\ref{imple-2-1}) can not be directly computed because we do not
have access to $\boldsymbol{A}^T$. Nevertheless, we can unfold
$\boldsymbol{H} \boldsymbol{\lambda}^{k}$ and
$\boldsymbol{A}^T\boldsymbol{\beta}^k$ to obtain
\begin{align}
&\boldsymbol{H}\boldsymbol{\lambda}^{k}=\boldsymbol{H}
(\boldsymbol{\lambda}^{k-1}+\alpha(\bar{\boldsymbol{\lambda}}^{k}-
\boldsymbol{\lambda}^{k-1}))=\boldsymbol{H}(\boldsymbol{\lambda}^{k-1}
+\alpha\sigma_1(\boldsymbol{x}^{k}
\nonumber\\
&\qquad\quad-\boldsymbol{H}^T\boldsymbol{y}^{k}))
=\textstyle\boldsymbol{H}(\boldsymbol{\lambda}^{1}+\alpha\sigma_1
\sum_{j=2}^{k}(\boldsymbol{x}^{j}-\boldsymbol{H}^T\boldsymbol{y}^{j}))
\nonumber\\
&\qquad\qquad\qquad\qquad\overset{(a)}{=}\textstyle\alpha\sigma_1
\sum_{j=2}^{k}\boldsymbol{H}(\boldsymbol{x}^{j}-\boldsymbol{H}^T
\boldsymbol{y}^{j}),
\nonumber\\
&\boldsymbol{A}^T\boldsymbol{\beta}^{k}=\textstyle
\boldsymbol{A}^T(\boldsymbol{\beta}^{k-1}+\sigma_2
\boldsymbol{A}\boldsymbol{y}^{k})=\boldsymbol{A}^T(\boldsymbol{\beta}^{1}+
\sum_{j=2}^k \sigma_2\boldsymbol{A}\boldsymbol{y}^{j}))
\nonumber\\
&\overset{(b)}{=}\textstyle\sum_{j=2}^k \sigma_2\boldsymbol{A}^T
\boldsymbol{A}\boldsymbol{y}^{j}, \label{imple-2-2}
\end{align}
where $(a)$ and $(b)$ are obtained by setting
$\boldsymbol{\lambda}^{1} =\boldsymbol{0}$ and
$\boldsymbol{\beta}^{1}=\boldsymbol{0}$, respectively.
Substituting (\ref{imple-2-2}) into (\ref{imple-2-1}) yields
\begin{align}
&\boldsymbol{y}^{k+1}=\textstyle(\alpha\sigma_1\boldsymbol{H}
\boldsymbol{H}^T+\sigma_2\boldsymbol{D})^{-1}\Big(\alpha\sigma_1
\boldsymbol{H}\big(\boldsymbol{x}^{k+1}+\sum_{j=2}^{k}\boldsymbol{H}
\nonumber\\
&(\boldsymbol{x}^{j}-\boldsymbol{H}^T\boldsymbol{y}^{j})\big)
- \textstyle\sum_{j=2}^k\sigma_2\boldsymbol{A}^T\boldsymbol{A}
\boldsymbol{y}^{j}+\sigma_2\big(\boldsymbol{D}-
\boldsymbol{A}^T\boldsymbol{A}\big)\boldsymbol{y}^k\Big)
\label{imple-2-3}
\end{align}
Note that $\boldsymbol{H}\boldsymbol{H}^T$ is a diagonal matrix.
Also, recall that
$\boldsymbol{H}\boldsymbol{x}^{k+1}=[\boldsymbol{H}_1
\boldsymbol{x}_1^{k+1};\cdots;\boldsymbol{H}_l\boldsymbol{x}_l^{k+1}]$,
the vector $\boldsymbol{H}_i\boldsymbol{x}_i^{k+1}$ can be
obtained by the $i$th ES through the model parameter upload step.
Meanwhile, $\boldsymbol{A}^T\boldsymbol{A} \boldsymbol{y}^k$ only
involves information exchange among neighboring ESs. Therefore by
letting each ES sending its local variable
$\boldsymbol{y}_{l}^{k}$ to its neighboring ESs,
$\boldsymbol{y}_{l}^{k+1}$ can be easily calculated at each ES.

It should be mentioned that if $k+1=\bar{k}$, then the
$\boldsymbol{y}^{k+1}$-subproblem can not be solved in a
decentralized manner. Nevertheless, this is inconsequential
because we only need to acquire $\boldsymbol{x}^{\bar{k}}$ in the
last iteration. The $\boldsymbol{y}^{\bar{k}}$-subproblem listed
in Algorithm \ref{alg:1} is only for an analysis purpose.

\subsubsection{$\boldsymbol{\lambda}_{ij}^{k+1}$-update}
Observe that the update of $\boldsymbol{\lambda}_{ij}^{k+1}$ in
(\ref{sec2-1}) is replaced by a two-step update. In the first
step, we calculate $\bar{\boldsymbol{\lambda}}_{ij}^{k+1}$ in a
way similar to (\ref{sec2-1}). Afterwards, an over-relaxation
step, i.e.,
$\boldsymbol{\lambda}_{ij}^{k+1}=\boldsymbol{\lambda}_{ij}^k+
\alpha(\bar{\boldsymbol{\lambda}}_{ij}^{k+1}-\boldsymbol{\lambda}_{ij}^k)$,
is conducted to obtain $\boldsymbol{\lambda}_{ij}^{k+1}$. Breaking
the standard update into such a two-step procedure is essential to
the global convergence of the proposed algorithm. Here are some
intuitions. At the $(k+1)$th iteration, only a subset of users are
selected to update $\boldsymbol{x}_{ij}^{k+1}$. However, all
users, including those are selected or unselected, are
required to update $\boldsymbol{\lambda}_{ij}^{k+1}$. Thus there
exists an imbalance between the update of the primal and that of
the dual variables. To guarantee the convergence of the algorithm,
an over-relaxation step with an inertia of $\alpha$ is included to
constrain the speed of the dual update since the over-relaxation
step forces $\boldsymbol{\lambda}_{ij}^{k+1}$ to be close to
$\boldsymbol{\lambda}_{ij}^{k}$.


\section{Convergence Analysis}
\label{sec-convergence}
In this section, we provide a theoretical justification for our
proposed ADMM algorithm. Our main results are summarized as
follows.


\newtheorem{theorem}{Theorem}
\begin{theorem}
\label{theorem-1} Denote
$\{\boldsymbol{x}_i^*,\boldsymbol{y}_i^*\}_{i=1}^l$ as the optimal
solution to the problem (\ref{sec1-4}), where
$\boldsymbol{x}_i^{*}\triangleq
[\boldsymbol{x}_{i1}^*;\boldsymbol{x}_{i2}^*;\cdots;\boldsymbol{x}_{i|S_i|}^*]$.
At each iteration each user is selected/activated with probability
$\alpha$. The maximum number of iterations is set to $\bar{k}$. In
addition, it is assumed that $f_{ij}$ is $\mu$-strongly convex
(see (\ref{def-strong})) and the
$\boldsymbol{x}_{ij}^{k+1}$-subproblem is solved up to an
$\epsilon^{k+1}$-accuracy. If $\boldsymbol{P}$ is chosen such that
\begin{align}
\boldsymbol{P}\succcurlyeq \textstyle(\frac{1}{\alpha^2}-1)
\frac{\sigma_1}{\sigma_2}\boldsymbol{H}\boldsymbol{H}^T-\frac{\alpha}{4}
\boldsymbol{A}^T\boldsymbol{A},
\label{the-main1-1}
\end{align}
then the sequence generated by Algorithm \ref{alg:1} satisfies
\begin{align}
\textstyle\mathbb{E}\big[\big|\sum\limits_{i=1}^l
(f_i(\boldsymbol{x}_{avg,i}^{\bar{k}})-f_i(\boldsymbol{x}_i^{*}))\big|\big]
\leq & \textstyle\frac{\tilde{C}^0}{1+\alpha(\bar{k}-1)}+
\frac{\sum\limits_{t=1}^{\bar{k}}(\epsilon^t)^2\sum\limits_{i=1}^l
|S_i|}{2\mu\bar{k}}, \label{the-mainin1}
\end{align}
and
\begin{align}
&\textstyle\mathbb{E}\big[\sum_{i=1}^l\|\boldsymbol{x}_{avg,i}^{\bar{k}}
-\boldsymbol{H}_i^T\boldsymbol{y}_{avg,i}^{\bar{k}}\|_2+\|\boldsymbol{A}
\boldsymbol{y}_{avg}^{\bar{k}}\|_2\big]
\nonumber\\
\leq & \textstyle\frac{\tilde{C}^0}{\psi(1+\alpha(\bar{k}-1))}+
\frac{\sum_{t=1}^{\bar{k}}(\epsilon^t)^2\sum_{i=1}^l|S_i|}{2\mu\psi\bar{k}},
\label{the-mainin2}
\end{align}
\end{theorem}
where the expectation is taken over all possible realizations due
to the random user selection, and
\begin{align}
&\boldsymbol{x}_{avg,i}^{\bar{k}}\triangleq \textstyle\sum_{t=1}^{\bar{k}}
\delta^t\boldsymbol{x}_i^{t}, \
\boldsymbol{y}_{avg,i}^{\bar{k}}\triangleq \sum_{t=1}^{\bar{k}} \delta^t
\boldsymbol{y}_i^{t},
\nonumber\\
&\delta^{\bar{k}}=(1+\alpha(\bar{k}-1))^{-1}, \ \delta^t=\alpha\delta^{\bar{k}},
\ 1\leq t\leq\bar{k}-1,
\nonumber\\
&\psi\triangleq\min\left\{\{\|\boldsymbol{\lambda}_i^*\|_2
+\xi\}_{i=1}^l, \|\boldsymbol{\beta}^*\|_2+\xi\right\},
\end{align}
in which $\boldsymbol{\lambda}_i^*$ and $\boldsymbol{\beta}^*$ are
the optimal dual variables, $\xi$ is a small positive scalar and
$\tilde{C}^0$ is a constant.



\subsection{Discussions}
\label{sec-3} Note that the first term on the left-hand side of
(\ref{the-mainin2}), i.e.
$\|\boldsymbol{x}_{avg,i}^{\bar{k}}-\boldsymbol{H}_i^T
\boldsymbol{y}_{avg,i}^{\bar{k}}\|_2$, measures the discrepancy
between the $i$th ES's local variable and the local variables of
its serving users. The second term, i.e.
$\|\boldsymbol{A}\boldsymbol{y}_{avg}^{\bar{k}}\|_2$, measures the
discrepancy between different ESs' local variables. If the sum of
these two quantities is zero, it means that the proposed algorithm
achieves a consensus in which all nodes' (including ESs and
users) local model parameters are equal to each other.

To gain insight into our result, we now turn to the terms on the
right-hand side of (\ref{the-mainin1}) and (\ref{the-mainin2}). We
see that the first term approaches $0$ as $\bar{k}$ increases. The
second term is an error term which is dependent on $\epsilon^{t}$.
Suppose we set $\epsilon^{t}=0,\forall t$, which means that the
$\boldsymbol{x}_{ij}^{t}$-subproblem is solved exactly. In this
case, the second term vanishes and our proposed algorithm will
eventually achieve consensus and obtain the optimal solution as
$\bar{k}\rightarrow\infty$.



Nevertheless, in practice, it may be computationally expensive to
find the exact solution of the
$\boldsymbol{x}_{ij}^{t}$-subproblem. Consider the case where
$\{\epsilon^t\}$ is a non-zero sequence. If $\epsilon^t$ is fixed
as a constant scalar, say $\epsilon$, then the second term on the
right-hand side of (\ref{the-mainin1}) and (\ref{the-mainin2}) is
a function of $\mu$ and $\epsilon$. Recall that the value $\mu$ is
used to quantify the curviness of $f_{ij}$. Specifically, a larger
$\mu$ indicates a more curvy $f_{ij}$, and for a fixed $\epsilon$,
a more curvy function $f_{ij}$ means that
$\boldsymbol{x}_{ij}^{k+1}$ is more close to the optimal solution
of the subproblem. Hence a larger $\mu$ results in a smaller
error. Although the second term on the right-hand side of
(\ref{the-mainin1}) and (\ref{the-mainin2}) cannot be removed for
a nonzero $\epsilon$, our simulation results show that for a
reasonable value of $\epsilon$, our proposed algorithm can achieve
an accurate solution close enough to the optimal one. Instead
of choosing a fixed $\epsilon$, an alternative is to employ a
sequence $\{\epsilon^t\}$ with decreasing values of $\epsilon^t$.
One option is to let $\{(\epsilon^t)^2\}_{t=1}^{+\infty}$ be a
summable sequence, say $(\epsilon^t)^2=t^{-2}$. For such a choice,
$\sum_{t=1}^{\infty} (\epsilon^t)^2$ is a finite number and thus
the error term in (\ref{the-mainin1}) and (\ref{the-mainin2})
tends to $0$ as $\bar{k}$ increases.


We now discuss the design of the matrix $\boldsymbol{P}$. As
discussed in (\ref{imple-2}), in order to achieve decentralized
implementation, $\boldsymbol{P}$ should satisfy
$\alpha^{-1}\boldsymbol{P}=\boldsymbol{D}-
\boldsymbol{A}^T\boldsymbol{A}$, where $\boldsymbol{D}$ is a
diagonal matrix. Moreover, as stated in Theorem \ref{theorem-1},
$\boldsymbol{P}$ should also satisfy the condition
(\ref{the-main1-1}). Combining these two conditions leads to
\begin{align}
\boldsymbol{D}\succcurlyeq \textstyle\frac{1}{\alpha}\left(\frac{1}{\alpha^2}-1
\right)\frac{\sigma_1}{\sigma_2}\boldsymbol{H}\boldsymbol{H}^T+\frac{3}{4}
\boldsymbol{A}^T\boldsymbol{A}.
\label{the-p-1}
\end{align}
To satisfy the above condition, we write $\boldsymbol{D}$ as
$\boldsymbol{D}=\boldsymbol{D}_{in} \otimes\boldsymbol{I}_n$,
where $\boldsymbol{D}_{in}$ is an $l\times l$ diagonal matrix.
Note that $\boldsymbol{H}=\boldsymbol{H}_{dig}\otimes
\boldsymbol{I}_n$, where
$\boldsymbol{H}_{dig}\triangleq\text{blkdig}
\{\boldsymbol{h}_1;\cdots;\boldsymbol{h}_l\}$ and
$\boldsymbol{h}_i$ is an all-one row vector of size $|S_i|$. Also,
we have
$\boldsymbol{A}\triangleq\boldsymbol{A}_{in}\otimes\boldsymbol{I}_n$.
Thus (\ref{the-p-1}) can be equivalently written as
\begin{align}
&\textstyle\left(\boldsymbol{D}_{in}- \frac{1}{\alpha}\left(\frac{1}{\alpha^2}-1
\right)\frac{\sigma_1}{\sigma_2}\boldsymbol{H}_{dig}\boldsymbol{H}_{dig}^T-
\frac{3}{4}\boldsymbol{A}_{in}^T\boldsymbol{A}_{in}\right)\otimes \boldsymbol{I}_n
\succcurlyeq \boldsymbol{0}
\nonumber\\
&\Leftrightarrow \textstyle\boldsymbol{D}_{in}- \frac{1}{\alpha}\left(\frac{1}
{\alpha^2}-1\right)\frac{\sigma_1}{\sigma_2}\boldsymbol{H}_{dig}\boldsymbol{H}_{dig}^T
-\frac{3}{4}\boldsymbol{A}_{in}^T\boldsymbol{A}_{in}\succcurlyeq \boldsymbol{0}
\end{align}
Observe that
$\boldsymbol{H}_{dig}\boldsymbol{H}_{dig}^T\in\mathbb{R}^{l\times
l}$ is a diagonal matrix with its $i$th diagonal element being
$|S_i|$. On the other hand, since $\boldsymbol{A}_{in}$ is the
incidence matrix of the graph $G$,
$\boldsymbol{A}_{in}^T\boldsymbol{A}_{in}$ is the Laplacian matrix
of the graph $G$. Let $\boldsymbol{D}_{L}$ be a diagonal matrix
whose diagonal elements equal those of $\boldsymbol{A}_{in}^T
\boldsymbol{A}_{in}$. We have $2\boldsymbol{D}_L\succcurlyeq
\boldsymbol{A}_{in}^T\boldsymbol{A}_{in}$. Hence it can be readily
verified that the matrix $\boldsymbol{D}$ defined as
\begin{align}
\textstyle\boldsymbol{D}=\left(\frac{1}{\alpha}\left(\frac{1}{\alpha^2}-1\right)
\frac{\sigma_1}{\sigma_2}\boldsymbol{H}_{dig}\boldsymbol{H}_{dig}^T+\frac{3}{2}
\boldsymbol{D}_L\right)\otimes\boldsymbol{I}_n
\end{align}
satisfies the condition (\ref{the-p-1}).

In the following, we provide a proof of Theorem \ref{theorem-1}.
We first define a function that will be frequently used:
\begin{align}
F^{t}_{(\boldsymbol{G},\boldsymbol{x})}\triangleq(\boldsymbol{x}^*-
\boldsymbol{x}^{t})^T\boldsymbol{G}(\boldsymbol{x}^{t}-\boldsymbol{x}^{t-1})
\label{def-Ft}
\end{align}
where $\boldsymbol{G}$ is an arbitrary positive semidefinite
matrix. Also, we introduce the following inequalities that will be
used in our proof. Regarding (\ref{sec1-4}), according to (2.1) in
\cite{HeYuan12}, we know that the following variational inequality
holds for $\forall \boldsymbol{x}_i, \boldsymbol{y}_i$:
\begin{align}
\textstyle\sum\limits_{i=1}^l
\big(f_i(\boldsymbol{x}_i)-f_i(\boldsymbol{x}_i^*)
+\langle\boldsymbol{\lambda}_i^*,\boldsymbol{x}_{i}
-\boldsymbol{H}_i^T\boldsymbol{y}_i\rangle\big)+\langle\boldsymbol{\beta}^*,
\boldsymbol{A}\boldsymbol{y}\rangle\geq 0, \label{sec2-2-3}
\end{align}
where $\boldsymbol{\lambda}_i^*$ and $\boldsymbol{\beta}^*$ are
the optimal dual variables. Employing the Cauchy-Schwarz
inequality, we further have
\begin{align}
&\textstyle\sum\limits_{i=1}^l\big(f_i(\boldsymbol{x}_i)-f_i(\boldsymbol{x}_i^*)+
\|\boldsymbol{\lambda}_i^*\|_2\|\boldsymbol{x}_{i}-\boldsymbol{H}_i^T
\boldsymbol{y}_i\|_2\big)+\|\boldsymbol{\beta}^*\|_2\|\boldsymbol{A}
\boldsymbol{y}\|_2.
\nonumber\\
&\geq 0 \label{sec2-2-4}
\end{align}


\section{Proof of Theorem \ref{theorem-1}}
\label{sec-proof}
The proof of Theorem \ref{theorem-1} consists of three parts. In
the first part, we establish an inequality (\ref{proof-14}). Then
in the second part, based on (\ref{proof-14}) we obtain an
inequality (\ref{proof-17}) that is close to our final results,
except that the values of $\{\boldsymbol{\lambda}_i\}$ and
$\boldsymbol{\beta}$ remain to be determined. At last, by
assigning appropriate values for $\{\boldsymbol{\lambda}_i\}$ and
$\boldsymbol{\beta}$ we obtain the desired results.

\subsection{Part I}
\label{sec-proof-part1}
\subsubsection{The $\boldsymbol{y}^{k+1}$-subproblem}
Invoking the notations in (\ref{sec3-2-1}), the $\boldsymbol{y}^{k+1}$-subproblem,
$k+1<\bar{k}$, can be compactly written as
\begin{align}
\boldsymbol{y}^{k+1}=\arg\min\limits_{\boldsymbol{y}} \ &\textstyle\frac{\alpha
\sigma_1}{2}\|\boldsymbol{x}^{k+1}-\boldsymbol{H}^T\boldsymbol{y}+
\frac{\boldsymbol{\lambda}^k}{\alpha\sigma_1}\|_2^2+
\nonumber\\
&\textstyle\frac{\sigma_2}{2}\|\boldsymbol{A}\boldsymbol{y}+
\frac{\boldsymbol{\beta}^{k}}{\sigma_2}\|_2^2
+\frac{\sigma_2}{2}\|\boldsymbol{y}-\boldsymbol{y}^k\|_{\alpha^{-1}
\boldsymbol{P}}^2.
\end{align}
Taking the gradient of the objective function and set it to $\boldsymbol{0}$
yields
\begin{align}
&\boldsymbol{0}=\textstyle \boldsymbol{H}(-\boldsymbol{\lambda}^k+\alpha\sigma_1
(\boldsymbol{H}^T\boldsymbol{y}^{k+1}-\boldsymbol{x}^{k+1}))+\boldsymbol{A}^T
(\boldsymbol{\beta}^k+\sigma_2\boldsymbol{A}\boldsymbol{y}^{k+1})
\nonumber\\
&\textstyle\qquad+\frac{\sigma_2}{\alpha}\boldsymbol{P}(\boldsymbol{y}^{k+1}-
\boldsymbol{y}^{k})
\nonumber\\
&\overset{(a)}{=}\textstyle \boldsymbol{H}(-\boldsymbol{\lambda}^{k}
-\alpha(\bar{\boldsymbol{\lambda}}^{k+1}-
\boldsymbol{\lambda}^{k}))+\boldsymbol{A}^T\boldsymbol{\beta}^{k+1}+
\frac{\sigma_2}{\alpha}\boldsymbol{P}(\boldsymbol{y}^{k+1}-\boldsymbol{y}^{k})
\nonumber\\
&\overset{(b)}{=}\textstyle-\boldsymbol{H}\boldsymbol{\lambda}^{k+1}
+\boldsymbol{A}^T\boldsymbol{\beta}^{k+1}+
\frac{\sigma_2}{\alpha}\boldsymbol{P}(\boldsymbol{y}^{k+1}-\boldsymbol{y}^{k}),
\label{appendix-1-1}
\end{align}
where $(a)$ is due to the update rule of $\bar{\boldsymbol{\lambda}}^{k+1}$
and $\boldsymbol{\beta}^{k+1}$, while $(b)$ is due to the update rule
of $\boldsymbol{\lambda}^{k+1}$. Analogously, if $k+1=\bar{k}$, it
holds
\begin{align}
\boldsymbol{0}=-\boldsymbol{H}\bar{\boldsymbol{\lambda}}^{\bar{k}}+\boldsymbol{A}^T
\boldsymbol{\beta}^{\bar{k}}+\sigma_2\boldsymbol{P}(\boldsymbol{y}^{\bar{k}}-
\boldsymbol{y}^{\bar{k}-1}).
\label{appendix-1-2}
\end{align}
Multiplying $\boldsymbol{\boldsymbol{y}}^{*}-\boldsymbol{y}^{k+1}$ (resp.
$\boldsymbol{\boldsymbol{y}}^{*}-\boldsymbol{y}^{\bar{k}}$) to both sides
of (\ref{appendix-1-1}) (resp. (\ref{appendix-1-2})) yields
\begin{align}
\boldsymbol{0}=&(\boldsymbol{y}^*-\boldsymbol{y}^{k+1})^T
\big(-\alpha\boldsymbol{H}
\boldsymbol{\lambda}^{k+1}+\alpha\boldsymbol{A}^T\boldsymbol{\beta}^{k+1}+
\nonumber\\
&\qquad\qquad\qquad\quad\sigma_2\boldsymbol{P}(\boldsymbol{y}^{k+1}-
\boldsymbol{y}^{k})\big), \ k+1< \bar{k},
\nonumber\\
\boldsymbol{0}=&(\boldsymbol{y}^*-\boldsymbol{y}^{\bar{k}})^T
\big(-\boldsymbol{H}\bar{\boldsymbol{\lambda}}^{\bar{k}}+\boldsymbol{A}^T
\boldsymbol{\beta}^{\bar{k}}+\sigma_2\boldsymbol{P}(\boldsymbol{y}^{\bar{k}}
-\boldsymbol{y}^{\bar{k}-1})\big),
\nonumber\\
& k+1=\bar{k}.
\label{proof-1}
\end{align}
Additionally, according to the $\boldsymbol{\beta}^{k+1}$-update in
(\ref{sec2-2-y}), we have
\begin{align}
&0=(\boldsymbol{\beta}-\boldsymbol{\beta}^{k+1})^T\big(\sigma_2^{-1}
(\boldsymbol{\beta}^{k+1}-\boldsymbol{\beta}^{k})-\boldsymbol{A}
\boldsymbol{y}^{k+1}\big), \ k+1< \bar{k},
\nonumber\\
&0=(\boldsymbol{\beta}-\boldsymbol{\beta}^{\bar{k}})^T
\big(\alpha\sigma_2^{-1}(\boldsymbol{\beta}^{\bar{k}}-
\boldsymbol{\beta}^{\bar{k}-1})-\boldsymbol{A}\boldsymbol{y}^{\bar{k}}
\big), \ k+1=\bar{k}.
\label{proof-2}
\end{align}
where $\boldsymbol{\beta}$ is an arbitrary vector of the same
dimension as $\boldsymbol{\beta}^{k+1}$. Summing (\ref{proof-1})
and (\ref{proof-2}) yields
\begin{align}
&\textstyle0=V^{\bar{k}}+\alpha\sum_{t=1}^{\bar{k}-1} V^{t}
+\sigma_2\sum_{t=1}^{\bar{k}} F^{t}_{(\boldsymbol{P},\boldsymbol{y})}
\nonumber\\
&\qquad \textstyle+\alpha\sigma_2^{-1}\sum_{t=1}^{\bar{k}}
(\boldsymbol{\beta}-\boldsymbol{\beta}^{t})^T(\boldsymbol{\beta}^{t}-
\boldsymbol{\beta}^{t-1}),
\label{proof-2-2}
\end{align}
where $F^{t}_{(\boldsymbol{P},\boldsymbol{y})}$ is defined in (\ref{def-Ft})
and
\begin{align}
&V^{\bar{k}}\triangleq(\boldsymbol{y}^*-\boldsymbol{y}^{\bar{k}})^T
\big(-\boldsymbol{H}\bar{\boldsymbol{\lambda}}^{\bar{k}}+
\boldsymbol{A}^T\boldsymbol{\beta}^{\bar{k}}\big)
-(\boldsymbol{\beta}-\boldsymbol{\beta}^{\bar{k}})^T
\boldsymbol{A}\boldsymbol{y}^{\bar{k}},
\nonumber\\
&V^{t}\triangleq(\boldsymbol{y}^*-\boldsymbol{y}^{t})^T\big(-\boldsymbol{H}
\boldsymbol{\lambda}^{t}+\boldsymbol{A}^T\boldsymbol{\beta}^{t}\big)
-(\boldsymbol{\beta}-\boldsymbol{\beta}^{t})^T\boldsymbol{A}
\boldsymbol{y}^{t},
\nonumber\\
&\qquad \quad t<\bar{k}.
\end{align}

\subsubsection{The $\boldsymbol{x}^{k+1}$-subproblem}
Note that the $\boldsymbol{x}_{ij}^{k+1}$-subproblem is solved up
to an $\epsilon^{k+1}$ accuracy, which means that
\begin{align}
\|\boldsymbol{\tau}_{ij}^{k+1}\|_2 \leq \epsilon^{k+1} \quad
\forall j\in I_i^{k+1} \label{proof-3}
\end{align}
where
\begin{align}
\boldsymbol{\tau}_{ij}^{k+1}=\partial
f_{ij}(\boldsymbol{x}_{ij}^{k+1})+
\boldsymbol{\lambda}_{ij}^k+\sigma_1(\boldsymbol{x}_{ij}^{k+1}-
\boldsymbol{y}_{i}^k)
\label{proof-3-1}
\end{align}
Based on (\ref{proof-3}), we can arrive at the following
inequality (see Appendix \ref{sec-appendix-2})
\begin{align}
0\leq &\textstyle (\alpha-1)(F_i^k+M_i^k+G_i^k)+
\mathbb{E}_{\boldsymbol{a}_i^{k+1}}\Big[F_i^{k+1}+M_i^{k+1}+
\nonumber\\
&\textstyle(1-\alpha)G_i^{k+1}+T_i^{k+1}+\frac{\alpha|S_i|
(\epsilon^{k+1})^2}{2\mu}\big|\{\boldsymbol{a}_i^{t}\}\Big]
\label{proof-4}
\end{align}
where $\{\boldsymbol{a}_i^{t}\}$ is used to represent
$\{\{\boldsymbol{a}_i^{t} \}_{i=1}^{l}\}_{t=1}^{k}$,
$\boldsymbol{a}_i^{k+1}\triangleq
\hat{\boldsymbol{a}}_i^{k+1}\otimes \boldsymbol{1}_{n}$,
$\hat{\boldsymbol{a}}_i^{k+1}\in\mathbb{R}^{|S_i|}$ is a random
binary vector with its $j$th element $\hat{a}_{ij}^{k+1}$ equal to
$1$ if user $u_{ij}$ is selected, and $0$ if otherwise, and $F_i^k
\triangleq f_i(\boldsymbol{x}_i^{*})-f_i(\boldsymbol{x}_i^{k})$,
$M_i^k\triangleq(\boldsymbol{x}_{i}^{*}-\boldsymbol{x}_{i}^{k})^T
\boldsymbol{\lambda}_{i}^k$,
\begin{align}
&G_i^k\triangleq\sigma_1(\boldsymbol{x}_{i}^{*}-\boldsymbol{x}_{i}^{k})^T
(\boldsymbol{x}_{i}^{k}-\boldsymbol{H}_{i}^T\boldsymbol{y}_{i}^k),
\nonumber\\
&T_i^{k+1}\triangleq\sigma_1(\boldsymbol{x}_{i}^{*}-\boldsymbol{x}_{i}^{k+1})^T
\boldsymbol{H}_i^T(\boldsymbol{y}_{i}^{k+1}-\boldsymbol{y}_{i}^{k}).
\end{align}
Regarding the conditional expectation, we have
\begin{align}
&\textstyle \mathbb{E}_x[f(x,y)|y]\geq 0\Leftrightarrow \int
f(x,y)p(x|y)dx\geq 0
\nonumber\\
\Leftrightarrow &\textstyle \int p(y)\left(\int f(x,y)p(x|y)dx
\right)dy\geq 0
\nonumber\\
\Rightarrow &\textstyle \int f(x,y)p(x,y)dxdy=\mathbb{E}_{x,y}
[f(x,y)]\geq 0.
\end{align}
Applying the above formula to (\ref{proof-4}) and summing the
resulting inequalities for all $i$, we have
\begin{align}
&0\leq \textstyle \mathbb{E}_{\{\{\boldsymbol{a}_i^{t}\}_{i=1}^{l}
\}_{t=1}^{k+1}}\Big[\sum\limits_{i=1}^{l}\big((\alpha-1)
(F_i^k+M_i^k)+F_i^{k+1}+M_i^{k+1}
\nonumber\\
&\textstyle +(\alpha-1)(G_i^k-G_i^{k+1})+T_i^{k+1}+\frac{\alpha|S_i|
(\epsilon^{k+1})^2}{2\mu}\big)\Big], \ k+1\leq \bar{k},
\end{align}
Hereafter we omit the subscript in $\mathbb{E}$ for the sake of simplicity.
Summing the above inequality for all $k+1$s (up to $\bar{k}$) yields
\begin{align}
&0\overset{(a)}{\leq} \textstyle C_i^0+C_i^1+\mathbb{E}\big[F_i^{\bar{k}}+
M_i^{\bar{k}}+\alpha\textstyle\sum_{t=1}^{\bar{k}-1}(F_i^t+M_i^t)
\nonumber\\
&\textstyle\qquad\qquad\qquad\qquad+(1-\alpha)G_i^{\bar{k}}+
\sum_{t=1}^{\bar{k}}T_i^{t}\big]
\nonumber\\
&\overset{(b)}{=}\textstyle\underbrace{C_i^0+C_i^1+\mathbb{E}
\big[F_i^{\bar{k}}+(\boldsymbol{x}_{i}^{*}-\boldsymbol{x}_{i}^{\bar{k}})^T
\bar{\boldsymbol{\lambda}}_{i}^{\bar{k}}+\alpha\textstyle\sum_{t=1}^{\bar{k}-1}
(F_i^t+M_i^t)}_{\text{[(\ref{proof-5-1})-1]}}
\nonumber\\
&\qquad\qquad\qquad\qquad\underbrace{\textstyle+\sum_{t=1}^{\bar{k}}T_i^{t}
\big]}_{\text{[(\ref{proof-5-1})-1]}}
\nonumber\\
&\overset{(c)}{=}\textstyle\text{[(\ref{proof-5-1})-1]}+
(\bar{\boldsymbol{\lambda}}_i^{\bar{k}}-\boldsymbol{\lambda}_i)^T
\big(\underbrace{\textstyle\frac{1}{\sigma_1}(\boldsymbol{\lambda}_{i}^{\bar{k}-1}
-\bar{\boldsymbol{\lambda}}_{i}^{\bar{k}})+\boldsymbol{x}_{i}^{\bar{k}}-
\boldsymbol{H}_i^T\boldsymbol{y}_{i}^{\bar{k}}}_{\text{[(\ref{proof-5-1})-2]}}\big)
\nonumber\\
&\qquad\textstyle+\alpha\sum\limits_{t=1}^{\bar{k}-1}(\boldsymbol{\lambda}_i^{t}
-\boldsymbol{\lambda}_i)^T\big(\underbrace{\textstyle\frac{1}{\alpha\sigma_1}
(\boldsymbol{\lambda}_{i}^{t-1}-\boldsymbol{\lambda}_{i}^{t})+\boldsymbol{x}_{i}^{t}
-\boldsymbol{H}_i^T\boldsymbol{y}_{i}^{t}\big)}_{\text{[(\ref{proof-5-1})-3]}}
\nonumber\\
&\textstyle\overset{(d)}{=}C_i^0+C_i^1+\mathbb{E}\Big[A_i^{\bar{k}}+
\alpha\sum\limits_{t=1}^{\bar{k}-1}A_i^{t}+
\sum_{t=1}^{\bar{k}}T_i^{t}+
\nonumber\\
&\underbrace{\textstyle\frac{1}{\sigma_1}
(\bar{\boldsymbol{\lambda}}_i^{\bar{k}}-\boldsymbol{\lambda}_i)^T
(\boldsymbol{\lambda}_{i}^{\bar{k}-1}-\bar{\boldsymbol{\lambda}}_{i}^{\bar{k}})
+\frac{1}{\sigma_1}\sum\limits_{t=1}^{\bar{k}-1}(\boldsymbol{\lambda}_i^{t}-
\boldsymbol{\lambda}_i)^T(\boldsymbol{\lambda}_{i}^{t-1}-
\boldsymbol{\lambda}_{i}^{t})}_{\text{[(\ref{proof-5-1})-4]}}\Big],
\label{proof-5-1}
\end{align}
where $C_i^0\triangleq(\alpha-1)(F_i^0+M_i^0+G_i^0)$ and
$C_i^1\triangleq
\frac{\alpha|S_i|}{2\mu}\sum_{t=1}^{\bar{k}}(\epsilon^t)^2$ are
constants, $\boldsymbol{\lambda}_i$ in $(c)$ is an arbitrary
vector, and $A_i^{\bar{k}}$ and $A_i^t$ in $(d)$ are defined as
\begin{align}
&A_i^{\bar{k}}\triangleq F_i^{\bar{k}}+
(\boldsymbol{x}_i^*-\boldsymbol{x}_i^{\bar{k}})^T
\boldsymbol{\bar{\lambda}}_{i}^{\bar{k}}+
(\bar{\boldsymbol{\lambda}}_i^{\bar{k}}-\boldsymbol{\lambda}_i)^T
(\boldsymbol{x}_{i}^{\bar{k}}-\boldsymbol{H}_i^T\boldsymbol{y}_{i}^{\bar{k}}),
\nonumber\\
&A_i^t\triangleq F_i^{t}+M_i^t+(\boldsymbol{\lambda}_i^{t}-
\boldsymbol{\lambda}_i)^T(\boldsymbol{x}_{i}^{t}-
\boldsymbol{H}_i^T\boldsymbol{y}_{i}^{t}), \ t<\bar{k},
\label{def-Ait}
\end{align}
Note that in (\ref{proof-5-1}), $(a)$ is due to the elimination of
the repeated terms in the summation, $(b)$ has invoked the fact
that $M_i^{\bar{k}}
+(1-\alpha)G_i^{\bar{k}}=(\boldsymbol{x}_{i}^{*}-\boldsymbol{x}_{i}^{\bar{k}})^T
\bar{\boldsymbol{\lambda}}_{i}^{\bar{k}}$ (since
$\boldsymbol{\lambda}_i^{\bar{k}}
\overset{(\ref{sec2-2})}{=}\boldsymbol{\lambda}_{i}^{\bar{k}-1}+\alpha
(\bar{\boldsymbol{\lambda}}_{i}^{\bar{k}}-\boldsymbol{\lambda}_{i}^{\bar{k}-1})$
and $\sigma_1(\boldsymbol{x}_{i}^{\bar{k}}-\boldsymbol{H}_{i}^T
\boldsymbol{y}_{i}^{\bar{k}})\overset{(\ref{sec2-2})}{=}
\bar{\boldsymbol{\lambda}}_{i}^{\bar{k}}-\boldsymbol{\lambda}_i^{\bar{k}-1}$),
$(c)$ has used the fact that $\text{[(\ref{proof-5-1})-2]}
\overset{(\ref{sec2-2})}{=}0$ and $\text{[(\ref{proof-5-1})-3]}
\overset{(\ref{sec2-2})}{=}0$, and $(d)$ is a simple
reorganization of the terms.

\subsubsection{Combining}
Summing up (\ref{proof-5-1}) for all $i$, $1\leq i\leq l$, and then summing
the resulting inequality with (\ref{proof-2-2}) yields
\begin{align}
&\textstyle 0 \leq \sum_{i=1}^l\Big(\mathbb{E}\big[A_i^{\bar{k}}+
\alpha\sum_{t=1}^{\bar{k}-1}A_i^{t}+\sum_{t=1}^{\bar{k}}T_i^{t}+
\text{[(\ref{proof-5-1})-4]}\big]+C_i^0
\nonumber\\
&\qquad\qquad\quad +C_i^1\Big)+(\ref{proof-2-2})
\nonumber\\
&=\textstyle A^{\bar{k}}+\alpha \sum_{t=1}^{\bar{k}-1}A^{t}+\sum_{i=1}^l
C_i^1+R
\label{proof-10-3}
\end{align}
where $\textstyle A^{\bar{k}}\triangleq\mathbb{E}[V^{\bar{k}}+\sum_{i=1}^{l}
A_i^{\bar{k}}]$, $\textstyle A^{t}\triangleq\mathbb{E}[V^t+\sum_{i=1}^{l}
A_i^t]$, $t< \bar{k}$, and $R$ represents the rest of the terms. According
to the derivations attached in Appendix \ref{sec-appendix-3}, it holds
$R\leq \textstyle\tilde{C}^0\big(\{\boldsymbol{\lambda}_i\},\boldsymbol{\beta}
\big)$, where
\begin{align}
\tilde{C}^0\big(\{\boldsymbol{\lambda}_i\},\boldsymbol{\beta}
\big)\triangleq&\textstyle\Big(\sum\limits_{i=1}^l\frac{\sigma_1}{2}
\|\boldsymbol{H}_i^T(\boldsymbol{y}_{i}^{*}-\boldsymbol{y}_{i}^{0})\|_2^2
+\frac{1}{2\sigma_1}\|\boldsymbol{\lambda}_i^{0}-\boldsymbol{\lambda}_i\|_2^2
\Big)
\nonumber\\
&\textstyle+\frac{\alpha}{2\sigma_2}\|\boldsymbol{\beta}-\boldsymbol{\beta}^{0}
\|_2^2+\frac{\alpha\sigma_2}{4}\|\boldsymbol{A}\boldsymbol{y}^{0}\|_2^2+
\sum_{i=1}^lC_i^0
\end{align}
is a function of $\{\boldsymbol{\lambda}_i\}$ and $\boldsymbol{\beta}$.
As such, (\ref{proof-10-3}) implies that
\begin{align}
0\leq \textstyle A^{\bar{k}}+\alpha \sum_{t=1}^{\bar{k}-1}A^{t}+\sum_{i=1}^l
C_i^1+\tilde{C}^0\big(\{\boldsymbol{\lambda}_i\},\boldsymbol{\beta}
\big)
\label{proof-14}
\end{align}

\subsection{Part II}
Eliminating the repeated terms in $A^t$ (also using the fact that
$\boldsymbol{x}_{i}^*
-\boldsymbol{H}_i^T\boldsymbol{y}_{i}^*=\boldsymbol{0}$ and
$\boldsymbol{A} \boldsymbol{y}^*=\boldsymbol{0}$), it can be
derived that
\begin{align}
A^{t}=\mathbb{E}\big[&\textstyle\sum_{i=1}^l\big(f_i(\boldsymbol{x}_i^*)
-f_i(\boldsymbol{x}_i^{t})-\langle\boldsymbol{\lambda}_i,\boldsymbol{x}_i^{t}-
\boldsymbol{H}_i^T\boldsymbol{y}_i^{t}\rangle\big)-
\nonumber\\
&\langle\boldsymbol{\beta},\boldsymbol{A}\boldsymbol{y}^{t}
\rangle\big], \ 1\leq t\leq \bar{k}.
\label{proof-16}
\end{align}
Substituting the right hand side of (\ref{proof-16}) into (\ref{proof-14}) yields
\begin{align}
&\textstyle\sum\nolimits_{i=1}^lC_i^1+\tilde{C}^0\big(\{\boldsymbol{\lambda}_i\},
\boldsymbol{\beta}\big)\overset{(\ref{proof-14})}{\geq}\textstyle
-A^{\bar{k}}-\alpha\sum_{t=1}^{\bar{k}-1}A^{t}
\nonumber\\
\overset{(\ref{proof-16})}{=}&\mathbb{E}
\Big[\underbrace{\textstyle\sum\limits_{i=1}^l\textstyle-\Big(f_i(\boldsymbol{x}_i^*)-
f_i(\boldsymbol{x}_i^{\bar{k}})+\alpha\sum_{t=1}^{\bar{k}-1}
(f_i(\boldsymbol{x}_i^*)-f_i(\boldsymbol{x}_i^{t}))
\Big)}_{[\text{(\ref{proof-16-1})-1}]}
\nonumber\\
&\quad\textstyle+\sum_{i=1}^l\langle\boldsymbol{\lambda}_i,
\boldsymbol{x}_i^{\bar{k}}-\boldsymbol{H}_i^T\boldsymbol{y}_i^{\bar{k}}\rangle+
\langle\boldsymbol{\beta},\boldsymbol{A}
\boldsymbol{y}^{\bar{k}}\rangle+
\nonumber\\
&\quad\textstyle\alpha\big(\sum_{i=1}^l\sum_{t=1}^{\bar{k}-1}
\langle\boldsymbol{\lambda}_i,\boldsymbol{x}_i^{t}-\boldsymbol{H}_i^T
\boldsymbol{y}_i^{t}\rangle\big)+\alpha\sum_{t=1}^{\bar{k}-1}
\langle\boldsymbol{\beta},\boldsymbol{A}\boldsymbol{y}^{t}\rangle\Big]
\nonumber\\
\overset{(a)}{\geq}& (1+\alpha(\bar{k}-1))
\underbrace{\textstyle\mathbb{E}\Big[\sum_{i=1}^l
\big(f_i(\boldsymbol{x}_{avg,i}^{\bar{k}})-f_i(\boldsymbol{x}_i^{*})
\big)+}_{[\text{(\ref{proof-16-1})-2}]}
\nonumber\\
&\underbrace{\textstyle\sum_{i=1}^l\big(\langle\boldsymbol{\lambda}_i,
\boldsymbol{x}_{avg,i}^{\bar{k}}-\boldsymbol{H}_i^T\boldsymbol{y}_{avg,i}^{\bar{k}}
\rangle+\langle\boldsymbol{\beta},\boldsymbol{A}\boldsymbol{y}_{avg}^{\bar{k}}
\rangle\big)\Big]}_{[\text{(\ref{proof-16-1})-2}]}
\label{proof-16-1}
\end{align}
where $\textstyle \boldsymbol{x}_{avg,i}^{\bar{k}}
\triangleq \sum_{t=1}^{\bar{k}} \delta^t \boldsymbol{x}_i^{t}, \
\boldsymbol{y}_{avg,i}^{\bar{k}}\triangleq \sum_{t=1}^{\bar{k}} \delta^t
\boldsymbol{y}_i^{t}, \ \delta^{\bar{k}}=(1+\alpha(\bar{k}-1))^{-1}, \
\delta^t=\alpha\delta^{\bar{k}}$, $t<\bar{k}$, and $(a)$ has invoked
Jensen's inequality (\ref{sec2-2-1}) in the follow manner:
\begin{align}
\textstyle [\text{(\ref{proof-16-1})-1}]\geq
\textstyle(1+\alpha(\bar{k}-1)) (\sum_{i=1}^l
f_i(\boldsymbol{x}_{avg,i}^{\bar{k}})-f_i(\boldsymbol{x}_i^{*}))
\end{align}
Multiplying $(1+\alpha(\bar{k}-1))^{-1}$ to both sides of (\ref{proof-16-1})
leads to
\begin{align}
[\text{(\ref{proof-16-1})-2}]
\leq &\textstyle\frac{\tilde{C}^0\big(\{\boldsymbol{\lambda}_i\},
\boldsymbol{\beta}\big)}{1+\alpha(\bar{k}-1)}+\frac{1}{2\mu}
\frac{\sum_{t=1}^{\bar{k}}(\epsilon^t)^2\sum_{i=1}^l|S_i|}{\bar{k}}.
\label{proof-17}
\end{align}
where we have invoked the definition of $C_i^1$ that is given below
(\ref{proof-5-1}).

\subsection{Part III}
To obtain our final result, let $\boldsymbol{\lambda}_i$ and
$\boldsymbol{\beta}$ in (\ref{proof-17}) be chosen as
\begin{align}
&\textstyle\boldsymbol{\lambda}_i=2(\|\boldsymbol{\lambda}_i^*\|_2+\xi)
\cdot\frac{\boldsymbol{x}_{avg,i}^{\bar{k}}-\boldsymbol{H}_i^T
\boldsymbol{y}_{avg,i}^{\bar{k}}}{\|\boldsymbol{x}_{avg,i}^{\bar{k}}-
\boldsymbol{H}_i^T\boldsymbol{y}_{avg,i}^{\bar{k}}\|_2},
\nonumber\\
&\textstyle\boldsymbol{\beta}=2(\|\boldsymbol{\beta}^*\|_2+\xi)\cdot
\frac{\boldsymbol{A}\boldsymbol{y}_{avg}^{\bar{k}}}{\|\boldsymbol{A}
\boldsymbol{y}_{avg}^{\bar{k}}\|_2}.
\label{proof-18}
\end{align}
where $\xi$ is a positive scalar. Substituting (\ref{proof-18}) into
both sides of (\ref{proof-17}) yields
\begin{align}
&\mathbb{E}\Big[\underbrace{\textstyle\sum_{i=1}^l\big(
f_i(\boldsymbol{x}_{avg,i}^{\bar{k}})-f_i(\boldsymbol{x}_i^{*})
\big)+2(\|\boldsymbol{\beta}^*\|_2+\xi)
\|\boldsymbol{A}\boldsymbol{y}_{avg}^{\bar{k}}\|_2}_{\text{[(\ref{proof-19})-1}]}
\nonumber\\
&\quad+\underbrace{\textstyle 2\sum_{i=1}^l(\|\boldsymbol{\lambda}_i^*\|_2+\xi)
\|\boldsymbol{x}_{avg,i}^{\bar{k}}-\boldsymbol{H}_i^T
\boldsymbol{y}_{avg,i}^{\bar{k}}\|_2}_{\text{[(\ref{proof-19})-1}]}\Big]
\nonumber\\
&\leq\underbrace{\textstyle\frac{\tilde{C}^0}
{1+\alpha(\bar{k}-1)}+\frac{\sum_{t=1}^{\bar{k}}(\epsilon^t)^2
\sum_{i=1}^l|S_i|}{2\mu\bar{k}}}_{\text{[(\ref{proof-19})-2]}},
\label{proof-19}
\end{align}
where $\tilde{C}^0$ is an upper bound of $\tilde{C}^0\big(\{\boldsymbol{\lambda}_i\},
\boldsymbol{\beta}\big)$ (recall that $\tilde{C}^0\big(\{\boldsymbol{\lambda}_i\},
\boldsymbol{\beta}\big)$ is finite since $\boldsymbol{\lambda}_i$ and
$\boldsymbol{\beta}$ have finite length). Regarding [(\ref{proof-19})-1], we have
\begin{align}
&\text{[(\ref{proof-19})-1]}\overset{\text{naturally}}{\geq}
\textstyle\sum_{i=1}^l \big(f_i(\boldsymbol{x}_{avg,i}^{\bar{k}})-
f_i(\boldsymbol{x}_i^{*})\big),
\\
&\text{[(\ref{proof-19})-1]}\overset{(\ref{sec2-2-4})}{\geq}
-\textstyle\sum_{i=1}^l
\big(f_i(\boldsymbol{x}_{avg,i}^{\bar{k}})-
f_i(\boldsymbol{x}_i^{*})\big),
\end{align}
which means that
\begin{align}
\text{[(\ref{proof-19})-1]}\geq \big|\textstyle\sum_{i=1}^l
\big(f_i(\boldsymbol{x}_{avg,i}^{\bar{k}})-f_i(\boldsymbol{x}_i^{*})\big)\big|.
\label{proof-21-1}
\end{align}
Combining (\ref{proof-21-1}) and (\ref{proof-19}) yields
\begin{align}
\textstyle\mathbb{E}\big[\big|\sum_{i=1}^l
f_i(\boldsymbol{x}_{avg,i}^{\bar{k}})-f_i(\boldsymbol{x}_i^{*})\big|\big]
\leq& \text{[(\ref{proof-19})-2]}.
\label{proof-22}
\end{align}
Additionally, we have
\begin{align}
\text{[(\ref{proof-19})-1]}\overset{(\ref{sec2-2-4})}{\geq}&
\textstyle\sum_{i=1}^l(\|\boldsymbol{\lambda}_i^*\|_2+\xi)
\|\boldsymbol{x}_{avg,i}^{\bar{k}}-\boldsymbol{H}_i^T
\boldsymbol{y}_{avg,i}^{\bar{k}}\|_2+
\nonumber\\
&(\|\boldsymbol{\beta}^*\|_2+\xi)\|\boldsymbol{A}\boldsymbol{y}_{avg}^{\bar{k}}\|_2
\nonumber\\
\geq &
\textstyle\psi\left(\sum_{i=1}^l\|\boldsymbol{x}_{avg,i}^{\bar{k}}
-\boldsymbol{H}_i^T\boldsymbol{y}_{avg,i}^{\bar{k}}\|_2+\|\boldsymbol{A}
\boldsymbol{y}_{avg}^{\bar{k}}\|_2\right) \label{proof-23}
\end{align}
where $\psi\triangleq\min\left\{\{\|\boldsymbol{\lambda}_i^*\|_2+\xi\}_{i=1}^l,
\|\boldsymbol{\beta}^*\|_2+\xi\right\}$. Combining (\ref{proof-23}) and
(\ref{proof-19}) leads to
\begin{align}
&\textstyle\mathbb{E}\big[\sum_{i=1}^l\|\boldsymbol{x}_{avg,i}^{\bar{k}}
-\boldsymbol{H}_i^T\boldsymbol{y}_{avg,i}^{\bar{k}}\|_2+\|\boldsymbol{A}
\boldsymbol{y}_{avg}^{\bar{k}}\|_2\big]
\nonumber\\
\leq &\textstyle\psi^{-1}[\text{(\ref{proof-19})-2}].
\label{proof-24}
\end{align}
Note that (\ref{proof-22}) and (\ref{proof-24}) are exactly the
results in Theorem \ref{theorem-1}. Our proof is completed here.

\section{Simulation Results}
\label{sec-simulation} In this section, we provide simulation
results to illustrate the performance of the proposed ADMM
algorithm (abbreviated as CFL-ADMM). To demonstrate the efficiency
of the algorithm, we compare it with the GT-SAGA (gradient
tracking-stochastic average gradient) method \cite{XinKhan20} and
the D-SGD (decentralized stochastic gradient descent) method
\cite{KoloskovaStich19}. We first discuss the setup of our
experiments and the implementation details of respective
algorithms.


\begin{figure}[!t]
\centering
\includegraphics[width=5.0cm]{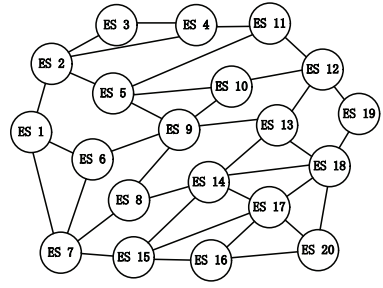}
\caption{Topology of the communication network of ESs. }
\label{fig1}
\end{figure}

\subsection{Setups}
\subsubsection{Experimental Setups} In our experiments,
the CFL network consists of $l=20$ ESs and $1000$ users. We assume
that each ES serves $S_i=50$ users. The communication network of
ESs is depicted in Fig. \ref{fig1}. We consider an
$\ell_2$-regularized logistic regression problem:
\begin{align}
\textstyle\mathop {\min }\limits_{\boldsymbol{x}\in\mathbb{R}^d} \
\sum_{i=1}^{l}\sum_{j=1}^{Si} f_{ij}(\boldsymbol{x}),
\label{simu-1}
\end{align}
where
$f_{ij}(\boldsymbol{x})=g_{ij}(\boldsymbol{x})+h_{ij}(\boldsymbol{x})$,
$g_{ij}(\boldsymbol{x})=\frac{\kappa}{2}\|\boldsymbol{x}\|_2^2$,
$\kappa=0.01$, and
\begin{align}
\textstyle h_{ij}(\boldsymbol{x})=\sum_{j'=1}^{n_{ij}}\Big(&-y_{ij,j'}
\cdot\text{log}\big((1+e^{-\boldsymbol{\omega}_{ij,j'}^T\boldsymbol{x}})^{-1}\big)-
\nonumber\\
&(1-y_{ij,j'})\cdot\text{log}\big(1-(1+e^{-\boldsymbol{\omega}_{ij,j'}^T
\boldsymbol{x}})^{-1}\big)\Big)
\label{simu-1-1}
\end{align}
in which
$\{\boldsymbol{\omega}_{ij,j'}\in\mathbb{R}^{n},y_{ij,j'}\in\{0,1\}\}$
is the $j'$th training sample stored at user $u_{ij}$. Note that
$f_{ij}$ is strongly convex and its gradient is Lipschitz
continuous.

Our experiments are based on the Credit 1
dataset\footnote{https://archive.ics.uci.edu/ml/datasets/default+of+credit+card+clients},
which consists of $30000$ real data samples. Each sample includes
24 entries, in which the first 23 entries along with a bias value
$1$ constitute $\boldsymbol{\omega}_{ij,j'}\in\mathbb{R}^{24}$ in
(\ref{simu-1-1}) and the last entry is the corresponding binary
label $y_{ij,j'}$. We randomly choose $20000$ samples for training
and each user is assigned with $20$ samples. For the proposed
CFL-ADMM, the $\boldsymbol{x}_{ij}^{k+1}$-subproblem is solved via
a simple gradient descent method. Since the gradient of the
objective function in the $\boldsymbol{x}_{ij}^{k+1}$-subproblem
is Lipschitz continuous, the gradient descent method is guaranteed
to converge to the optima, provided that the stepsize is
appropriately selected. The initial point of the gradient descent
method for solving the $\boldsymbol{x}_{ij}^{k+1}$-subproblem is
chosen to be the solution obtained in the last iteration, i.e.
$\boldsymbol{x}_{ij}^{k}$.



\subsubsection{Implementations of GT-SAGA and D-SGD}
Note that both GT-SAGA and D-SGD were originally developed for D2D
networks. Nevertheless, they can be easily adapted to the
considered CFL framework. Take GT-SAGA as an example. The GT-SAGA
aims to solve problems of the same form as (\ref{sec1-1}). In
GT-SAGA, it is assumed that each data-holder holds a local
objective function
$f_i(\boldsymbol{x})=\sum_{j=1}^{|S_i|}f_{ij}(\boldsymbol{x};
\mathcal{D}_{ij})$, where $\mathcal{D}_{ij}$ represents the data
set corresponding to the loss function $f_{ij}$. The GT-SAGA
assumes that there is no user and the data-holders collaboratively
solve (\ref{sec1-1}). In each iteration, each data-holder randomly
selects a portion of $f_{ij}$s to update the local model, followed
by an information exchange between the data-holders to enforce the
consensus among local variables. We can adapt the GT-SAGA to our
CFL framework by distributing $f_{ij}$ and $\mathcal{D}_{ij}$ to
user $u_{ij}$. In such a setting, each user first downloads the
model vector, say $\boldsymbol{y}_i^{k+1}$, from the ES, followed
by the computation of the gradient of $f_{ij}$ at
$\boldsymbol{y}_i^{k+1}$, and then uploads the gradient vector to
its associated ES for aggregation. The D-SGD method can be adapted
to our CFL framework in a similar way.

Note that when adapting those decentralized stochastic
gradient-based methods to the CFL framework, only a single
gradient descent step is allowed to be performed at each
iteration. Those methods which perform multiple rounds of gradient
descent at each iteration \cite{McMahanMoore17,LiuChen20,
LiuChen22,KarimireddyKale20,LiangShen19,YuanMa20,HaddadpourMahdavi19}
are not applicable. This is because for those decentralized
stochastic gradient-based methods, each user is required to upload
the gradient of $f_{ij}$ to its associated ES. More specifically,
suppose the user $u_{ij}$ receives a model parameter vector
$\boldsymbol{y}_i^{k+1}$ from the $i$th ES at the $(k+1)$th
iteration. Then the gradient of $f_{ij}$ should be computed at the
point $\boldsymbol{y}_i^{k+1}$. Performing multiple steps of
gradient descent at each user and then reporting the final
gradient will lead to incorrect results.

\begin{figure*}[!htbp]
    \centering
    \subfigure[\scriptsize{Error tolerance $\epsilon=1e-3$.}]
    {\includegraphics[width=5cm,height=4cm]{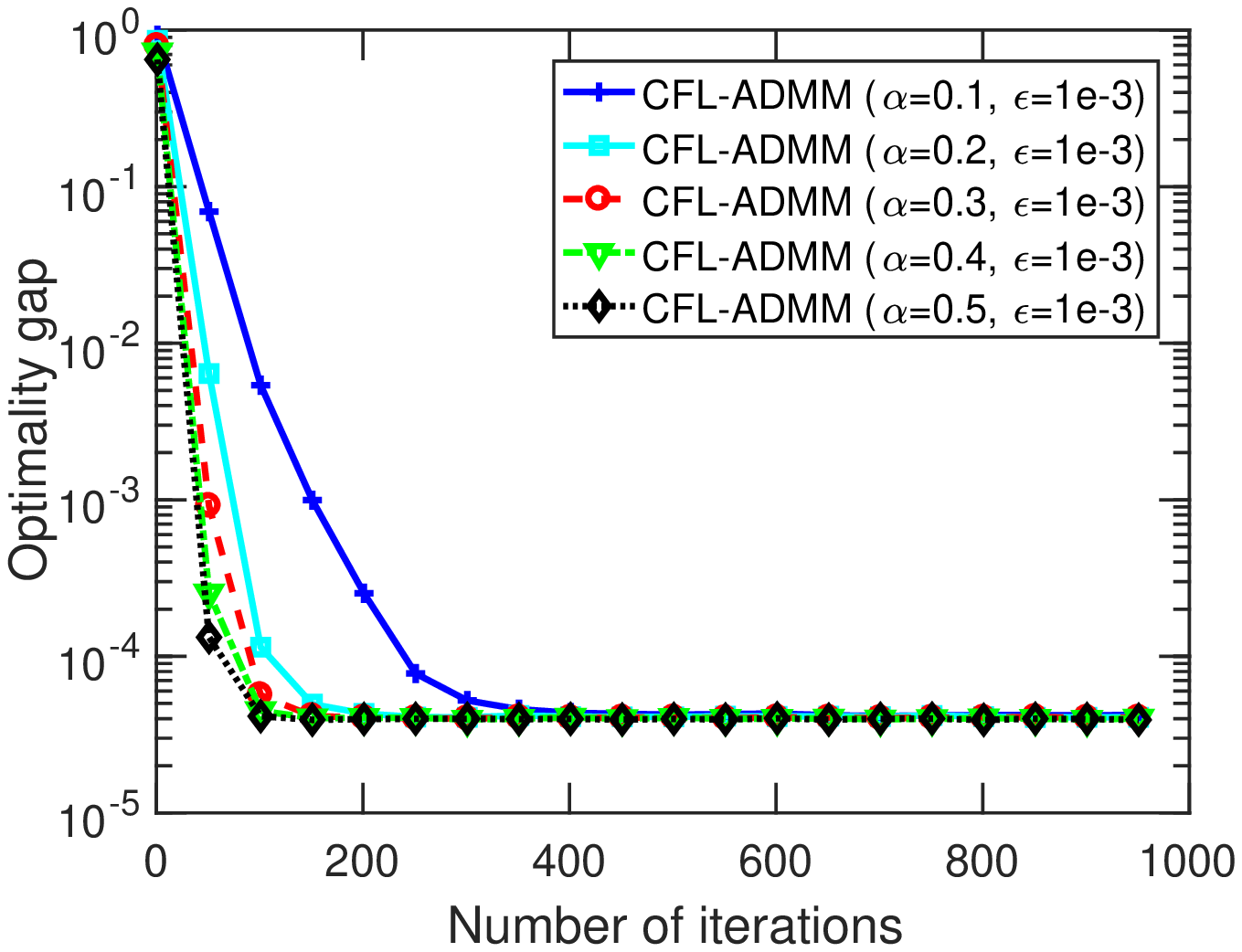}} \hfil
    \subfigure[\scriptsize{Error tolerance $\epsilon=1e-4$.}]
    {\includegraphics[width=5cm,height=4cm]{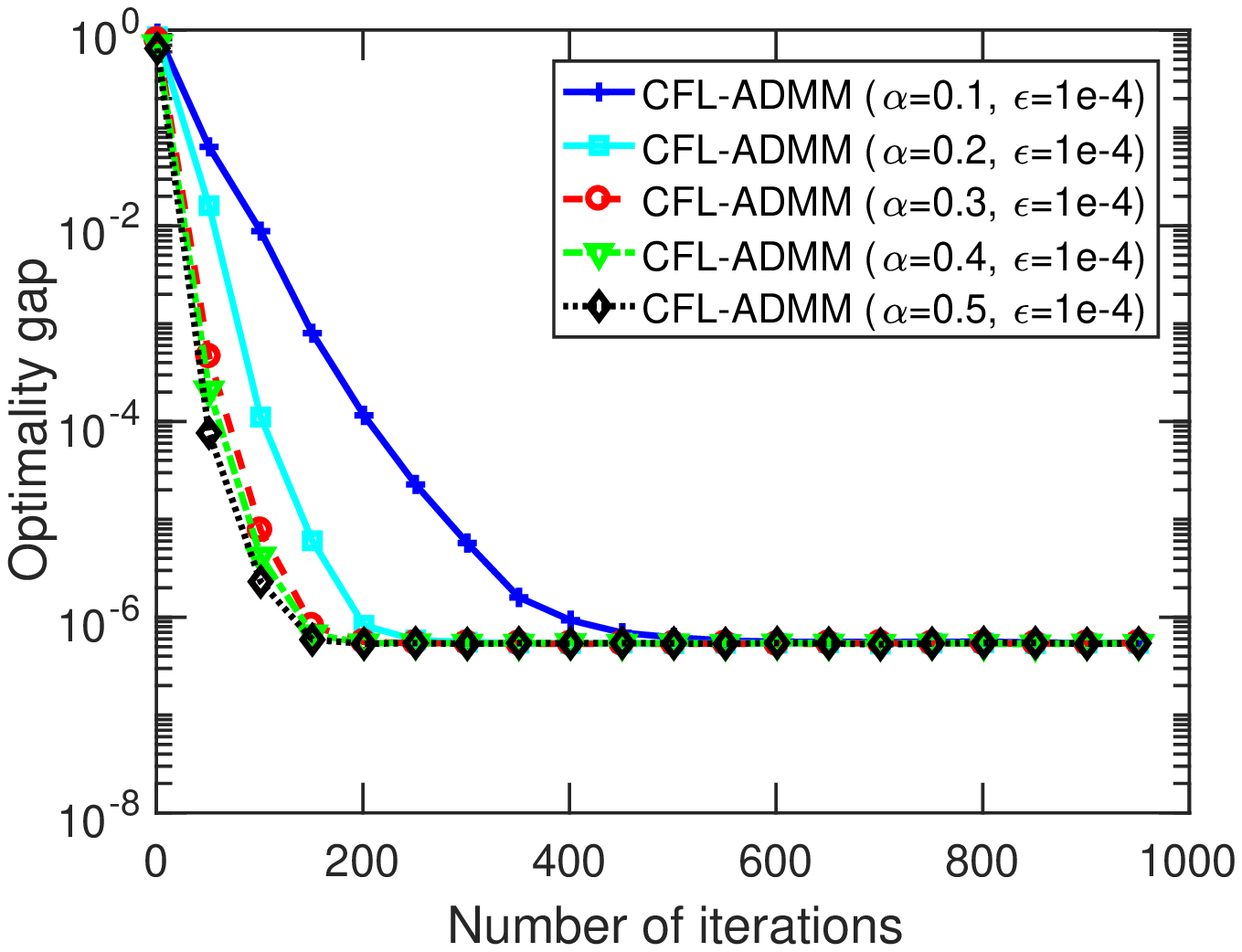}} \hfil
    \subfigure[\scriptsize{Error tolerance $\epsilon=1e-5$.}]
    {\includegraphics[width=5cm,height=4cm]{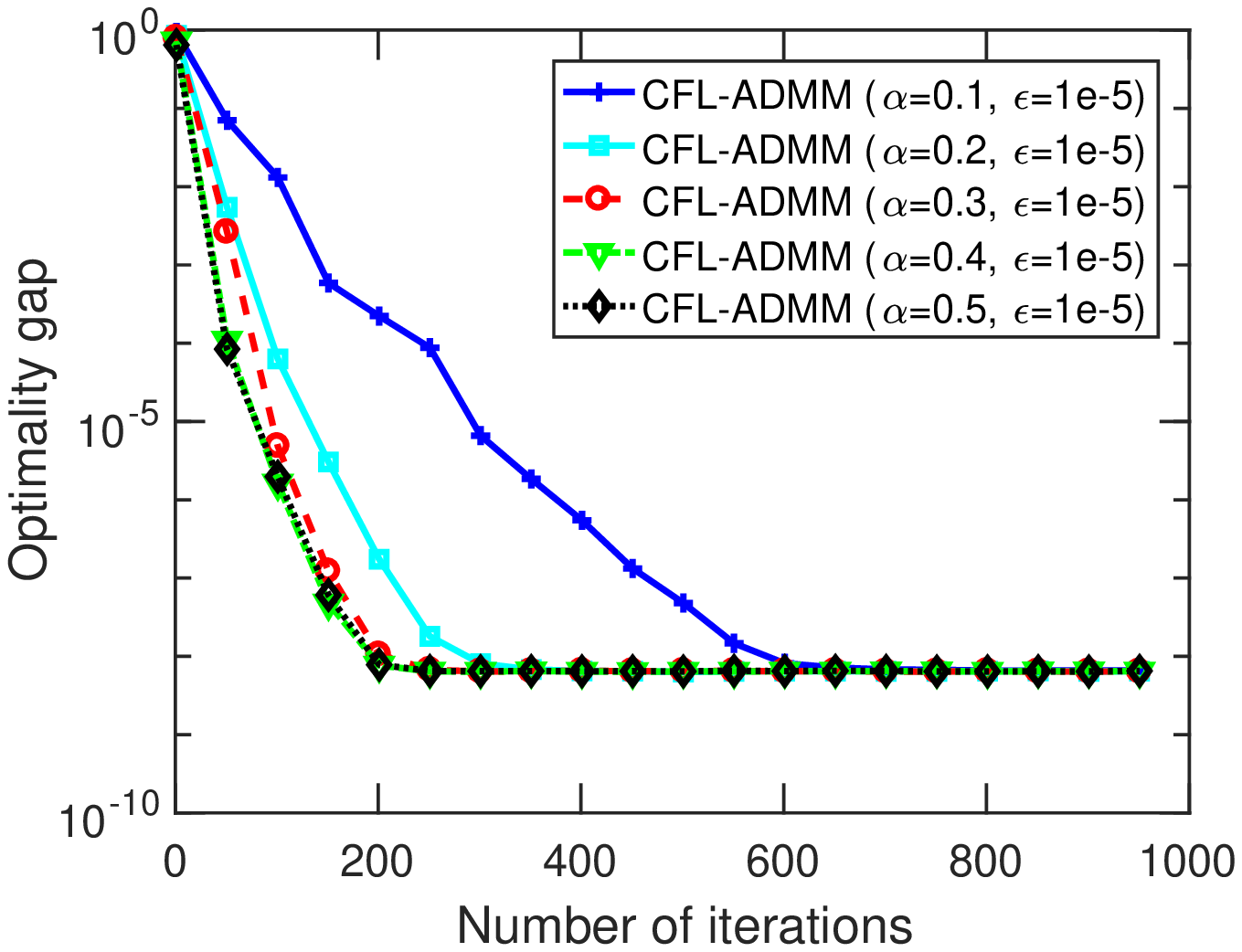}}
    \caption{$\ell_2$-regularized logistic regression: Optimality gap vs. the number of iterations.}
    \label{fig2}
\end{figure*}





\subsection{Results on $\ell_2$-Regularized Logistic Regression}
To evaluate the performance of the proposed method, the following
metric is introduced, namely, an optimality gap $d^k$ used to
measure the distance between the obtained solution and the optimal
solution:
\begin{align}
d^k\triangleq \frac{1}{\|\boldsymbol{x}^*\|_2^2\cdot
\sum_{i=1}^{l}|S_i|}\textstyle\sum\nolimits_{i=1}^{l}\sum\nolimits_{j=1}^{|S_i|}
\|\boldsymbol{x}_{ij}^k-\boldsymbol{x}^*\|_2^2,
 \label{simu-3}
\end{align}
where $\boldsymbol{x}_{ij}^k$ is the solution obtained at the
$k$th iteration, and $\boldsymbol{x}^*$ is the optimal solution of
the problem. Note that the \emph{optimality metric} is defined by
using the instantaneous output $\boldsymbol{x}_{ij}^k$ instead of
the time average defined in Theorem \ref{theorem-1}. This is
because the time average is overly pessimistic and leads to a
relatively slow convergence speed.


\begin{figure}[!t]
\centering
\includegraphics[width=6cm]{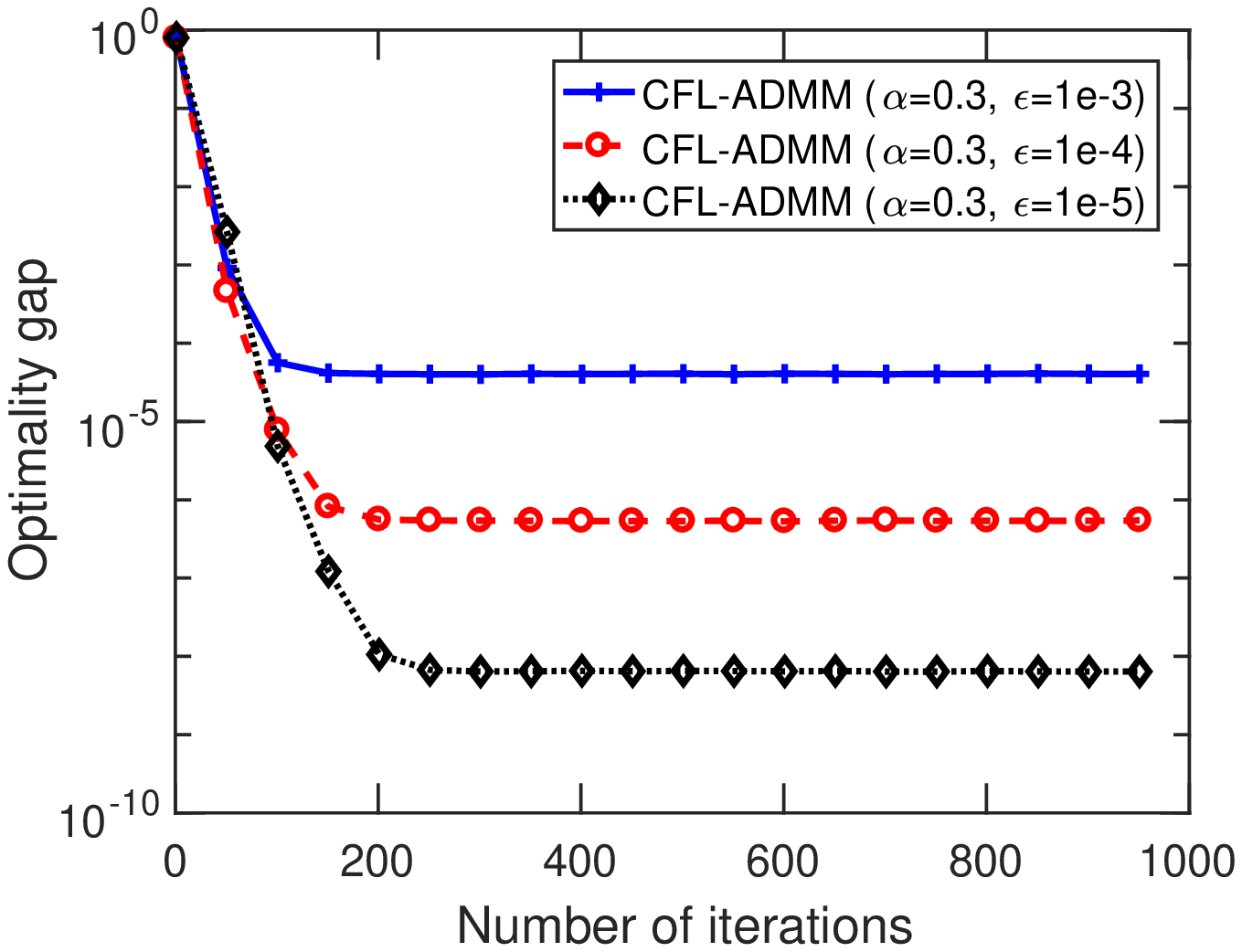}
\caption{$\ell_2$-regularized logistic regression: Optimality gap vs. the number
of iterations. }
\label{fig3}
\end{figure}


Fig. \ref{fig2} plots the optimality gap of the proposed CFL-ADMM
vs. the number of iterations under different selection
probabilities $\alpha$ and different error tolerances $\epsilon$.
Results are averaged over $100$ independent runs, with users
randomly selected for each run and each iteration. Clearly, when
using a nonzero $\epsilon$, the algorithm does not converge to the
true solution $\boldsymbol{x}^{*}$. Instead, it converges to a
neighborhood of $\boldsymbol{x}^{*}$. From Fig. \ref{fig2}, it can
be observed that the converged point is closer to
$\boldsymbol{x}^{*}$ when a smaller $\epsilon$ is employed. In
addition, it is observed that a larger user selection probability
$\alpha$ leads to a faster convergence speed. Nevertheless, the
performance improvement becomes insignificant as the selection
probability exceeds $\alpha>0.3$. Since the average amount of
communication overhead grows linearly with $\alpha$, it is better
to choose a moderate value of $\alpha$ to strike a reasonable
balance between the performance and the communication cost.

In Fig. \ref{fig3}, we evaluate the performance of the proposed
algorithm under different values of error tolerance $\epsilon$.
The user selection probability is set to $\alpha=0.3$. We see that
the choice of $\epsilon$ does not affect the convergence speed of
the proposed algorithm, which is in consistent with the results
reported in Theorem \ref{theorem-1}.




\begin{figure}[!t]
\centering
\includegraphics[width=6cm]{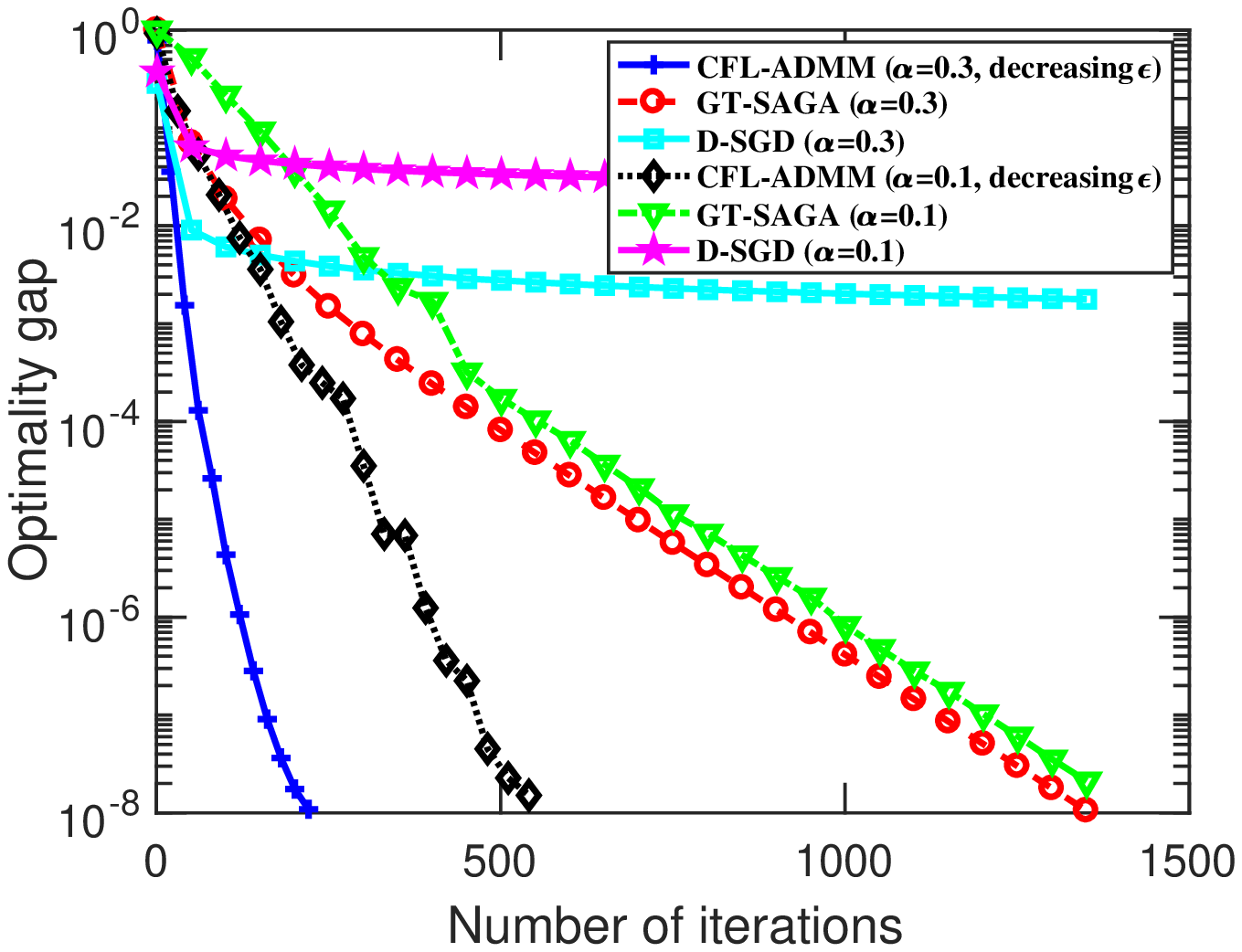}
\caption{$\ell_2$-regularized logistic regression: Optimality gap vs. the number of iterations. }
\label{fig6}
\end{figure}


Next, we compare the performance of our proposed algorithm with
GT-SAGA and D-SGD. The parameters of respective algorithms are
tuned to achieve the best performance. For our proposed algorithm,
instead of using a fixed $\epsilon$, we employ a decreasing error
tolerance sequence $\{\epsilon^{k}\}$ to ensure that it converges
to the optimal solution. More specifically, we set
$\epsilon^{k}=\frac{1}{100+k^2}$. Fig. \ref{fig6} plots the
optimality gap of respective algorithms vs. the number of
iterations. With a same $\alpha$, all three algorithms have the
same per-iteration communication cost. It can be observed that the
proposed CFL-ADMM converges much faster than the other two
stochastic gradient-based algorithms, which implies that the
proposed algorithm can attain a solution of a same quality with
much fewer rounds of communication, and thus achieves a higher
communication efficiency.

We would like to point out that the improved communication
efficiency of the proposed algorithm comes at the expense of
involving more computations at users. Specifically, for GT-SAGA
and D-SGD, each user only needs to compute the gradient of its
local objective function once at each iteration, while for the
proposed algorithm, each user needs to solve a subproblem up to a
certain accuracy, which usually requires several or tens of
iterations of gradient descent. Nevertheless, nowadays the
computing power of mobile devices such as smartphones has
increased to an impressive level. In contrast, as the information
are usually transmitted wirelessly from users to ESs,
communications are more expensive and power-consuming than
computations. In addition, more rounds of communications result in
a higher latency, which is also a critical factor that should be
considered in FL applications. In fact, since the initial point of
the $\boldsymbol{x}_{ij}^{k+1}$-subproblem of CFL-ADMM is chosen
as $\boldsymbol{x}_{ij}^{k}$, it only takes several iterations of
gradient descent (except for the first few tens of ADMM
iterations) to reach the specified accuracy. Therefore the
disadvantage of the proposed algorithm on the computational aspect
is not that significant.

\section{Conclusions}
\label{sec-conclusion} In this paper, we introduced a hybrid
centralized and decentralized FL framework (referred to as CFL) to
enhance the scalability of FL. The framework consists of multiple
servers, in which each server serves an individual set of devices
as in the conventional FL framework, and multiple servers form a
decentralized network. An ADMM algorithm was developed within such
a hybrid framework. The proposed ADMM randomly selects each user
with a certain probability at each iteration, thus alleviating the
heavy communication burden caused by the interaction between the
servers and the users. Moreover, the proposed ADMM allows the
subproblem to be inexactly solved at each user, making it amiable
for machine learning applications. Our theoretical analysis showed
that the proposed ADMM enjoys a $O(1/k)$ convergence rate.
Numerical results were provided to illustrate the effectiveness
and superiority of the proposed ADMM.

\appendices


\section{From (\ref{proof-3}) to (\ref{proof-4})}
\label{sec-appendix-2}
Since (\ref{proof-3-1}) holds for $\forall j\in I_i^{k+1}$, we
can compactly rewrite it as
\begin{align}
&\boldsymbol{a}_i^{k+1}\odot\boldsymbol{\tau}_i^{k+1}
\nonumber\\
=&\boldsymbol{a}_i^{k+1}\odot\big(\partial f_i
\big(\boldsymbol{x}_i^{k+1}\big)+\boldsymbol{\lambda}_i^k+
\sigma_1(\boldsymbol{x}_i^{k+1}-\boldsymbol{H}_i^T
\boldsymbol{y}_{i}^k)\big)
\label{appendix-2-1-1}
\end{align}
where $\boldsymbol{x}_i$, $\boldsymbol{\lambda}_i$,
$\boldsymbol{H}_i$ and $\boldsymbol{y}_{i}$ are defined in
(\ref{sec3-2-1}), $\odot$ represents element-wise product,
$\boldsymbol{\tau}_i\triangleq[\boldsymbol{\tau}_{i1};\cdots;
\boldsymbol{\tau}_{i|S_i|}]$, $\partial
f_i\big(\boldsymbol{x}_i)\triangleq [\partial
f_{i1}(\boldsymbol{x}_{i1});\cdots;\partial
f_{i|S_i|}(\boldsymbol{x}_{i|S_i|})]$,
$\boldsymbol{a}_i^{k+1}\triangleq
\hat{\boldsymbol{a}}_i^{k+1}\otimes \boldsymbol{1}_{n}$ and
$\hat{\boldsymbol{a}}_i^{k+1}\in\mathbb{R}^{|S_i|}$ is a random
binary vector with its $j$th element $\hat{a}_{ij}^{k+1}$ equal to
$1$ if user $u_{ij}$ is selected while equal to $0$ otherwise.
Note that the $j$th element of $\hat{\boldsymbol{a}}_i^{k+1}$ has
a probability of $\alpha$ (resp. $1-\alpha$) to be equal to $1$
(resp. $0$). Multiplying $\boldsymbol{a}_i^{k+1}
\odot(\boldsymbol{x}_{i}^*-\boldsymbol{x}_{i}^{k+1})$ to both
sides of (\ref{appendix-2-1-1}) and then taking the expectation of
the resulting equality yields
\begin{align}
&\mathbb{E}_{\boldsymbol{a}_i^{k+1}}\big[(\boldsymbol{a}_i^{k+1}\odot
(\boldsymbol{x}_{i}^*-\boldsymbol{x}_{i}^{k+1}))^T\boldsymbol{\tau}_{i}^{k+1}\big |
\{\boldsymbol{a}_i^{t}\}\big]
\nonumber\\
= &\mathbb{E}_{\boldsymbol{a}_i^{k+1}}\big[(\boldsymbol{a}_i^{k+1}\odot
(\boldsymbol{x}_{i}^*-\boldsymbol{x}_{i}^{k+1}))^T\big(\partial f_{i}
\big(\boldsymbol{x}_{i}^{k+1}\big)+\boldsymbol{\lambda}_{i}^k+
\nonumber\\
&\qquad\quad\sigma_1(\boldsymbol{x}_{i}^{k+1}-\boldsymbol{H}_{i}^T
\boldsymbol{y}_{i}^k)\big)\big | \{\boldsymbol{a}_i^{t}\}\big],
\label{appendix-2-1}
\end{align}
where $\{\boldsymbol{a}_i^{t}\}$ is used to represent
$\{\{\boldsymbol{a}_i^{t} \}_{i=1}^{l}\}_{t=1}^{k}$ and the above
equality comes from the fact that $(\boldsymbol{a}_i^{k+1}
\odot\boldsymbol{x})^T(\boldsymbol{a}_i^{k+1}\odot\boldsymbol{y})=
(\boldsymbol{a}_i^{k+1}\odot\boldsymbol{x})^T\boldsymbol{y}$,
$\forall \boldsymbol{x}$, $\boldsymbol{y}$. Clearly, taking an
expectation w.r.t. $\boldsymbol{a}_i^{k+1}$ is equivalent to
taking an expectation w.r.t. $\{\hat{a}_{ij}^{k+1}\}_j$. Note that
the expectation in (\ref{appendix-2-1}) is taken only w.r.t.
$\boldsymbol{a}_i^{k+1}$ instead of all random vectors because the
randomness of other random vectors, say
$\boldsymbol{x}_{i}^{k+1}$, originates in that of
$\boldsymbol{a}_i^{k+1}$. Next, we separately upper bound the
terms on the right hand side of (\ref{appendix-2-1}).

\subsubsection{Bounding the first term}
Consider the first term in (\ref{appendix-2-1}), we have
\begin{align}
&\mathbb{E}_{\boldsymbol{a}_i^{k+1}}\big[(\boldsymbol{a}_i^{k+1}\odot
(\boldsymbol{x}_{i}^*-\boldsymbol{x}_{i}^{k+1}))^T\partial f_{i}
(\boldsymbol{x}_{i}^{k+1})\big | \{\boldsymbol{a}_i^{t}\}\big]
\nonumber\\
\overset{(a)}{\leq}&\textstyle
\mathbb{E}_{\boldsymbol{a}_i^{k+1}}
\Big[\sum_{j=1}^{|S_i|}\hat{a}_{ij}^{k+1}\Big(f_{ij}(\boldsymbol{x}_{ij}^*)-
f_{ij}(\boldsymbol{x}_{ij}^k)+f_{ij}(\boldsymbol{x}_{ij}^k)
\nonumber\\
&\textstyle\quad\qquad-f_{ij}(\boldsymbol{x}_{ij}^{k+1})-\frac{\mu}{2}
\big\|\boldsymbol{x}_{ij}^*-\boldsymbol{x}_{ij}^{k+1}\big\|_2^2\Big)
\Big | \{\boldsymbol{a}_i^{t}\}\Big]
\nonumber\\
\overset{(b)}{=}&\textstyle \alpha \big(f_i(\boldsymbol{x}_i^*)-
f_i(\boldsymbol{x}_i^k)\big)+\mathbb{E}_{\boldsymbol{a}_i^{k+1}}
\big[\sum_{j=1}^{|S_i|}\hat{a}_{ij}^{k+1}\big(f_{ij}(\boldsymbol{x}_{ij}^{k})
\nonumber\\
&\textstyle-f_{ij}(\boldsymbol{x}_{ij}^{k+1})-\frac{\mu}{2}\big\|\boldsymbol{x}_{ij}^*-
\boldsymbol{x}_{ij}^{k+1}\big\|_2^2\big)\big | \{\boldsymbol{a}_i^{t}\}\big]
\nonumber\\
\overset{(c)}{=}&\textstyle (\alpha-1) \big(f_i(\boldsymbol{x}_i^*)-
f_i(\boldsymbol{x}_i^k)\big)+\mathbb{E}_{\boldsymbol{a}_i^{k+1}}
\big[f_i(\boldsymbol{x}_i^*)-f_i(\boldsymbol{x}_i^{k+1})
\nonumber\\
&\textstyle-\frac{\mu}{2}\|\boldsymbol{x}_{ij}^*-\boldsymbol{x}_{ij}^{k+1}\|_2^2
\big | \{\boldsymbol{a}_i^{t}\}\big]
\label{appendix-2-2}
\end{align}
where $(a)$ has invoked (\ref{def-strong}), $(b)$ is because
\begin{align}
&\textstyle\mathbb{E}_{\boldsymbol{a}_i^{k+1}}\big[\sum_{j=1}^{|S_i|}
\hat{a}_{ij}^{k+1}\big(f_{ij}(\boldsymbol{x}_{ij}^*)-f_{ij}
(\boldsymbol{x}_{ij}^k)\big)\big | \{\boldsymbol{a}_i^{t}\}\big]
\nonumber\\
=&\textstyle\alpha\big(f_i(\boldsymbol{x}_i^*)-f_i(\boldsymbol{x}_i^k)\big)
\end{align}
and $(c)$ is due to
\begin{align}
&\textstyle f_i(\boldsymbol{x}_i^*)-f_i(\boldsymbol{x}_i^k)+
\mathbb{E}_{\boldsymbol{a}_i^{k+1}}\Big[\sum\limits_{j=1}^{|S_i|}
\hat{a}_{ij}^{k+1}\big(f_{ij}(\boldsymbol{x}_{ij}^k)-
f_{ij}(\boldsymbol{x}_{ij}^{k+1})\big)\Big|
\nonumber\\
&\qquad\qquad\qquad\qquad\qquad\quad\{\boldsymbol{a}_i^{t}\}\Big]
\nonumber\\
&= f_i(\boldsymbol{x}_i^*)-f_i(\boldsymbol{x}_i^k)+
\mathbb{E}_{\boldsymbol{a}_i^{k+1}}\big[f_{i}(\boldsymbol{x}_{i}^{k})-
f_{i}(\boldsymbol{x}_{i}^{k+1})\big|\{\boldsymbol{a}_i^{t}\}\big]
\nonumber\\
&=\mathbb{E}_{\boldsymbol{a}_i^{k+1}}\big[f_i(\boldsymbol{x}_i^*)-
f_i(\boldsymbol{x}_i^{k+1})\big|\{\boldsymbol{a}_i^{t}\}\big],
\label{appendix-2-2-1}
\end{align}
in which the first equality is because $\boldsymbol{x}_{ij}^{k+1}=
\boldsymbol{x}_{ij}^k$ when $\hat{a}_{ij}^{k+1}=0$. Thus we have
\begin{align}
&\textstyle\sum_{j=1}^{|S_i|}
\hat{a}_{ij}^{k+1}\big(f_{ij}(\boldsymbol{x}_{ij}^k)-
f_{ij}(\boldsymbol{x}_{ij}^{k+1})\big)
\nonumber\\
=&\textstyle\sum_{j=1}^{|S_i|}
\big(f_{ij}(\boldsymbol{x}_{ij}^k)-
f_{ij}(\boldsymbol{x}_{ij}^{k+1})\big)
=f_{i}(\boldsymbol{x}_{i}^{k})- f_{i}(\boldsymbol{x}_{i}^{k+1})
\end{align}
Note that the expectation in the second line of
(\ref{appendix-2-2-1}) can not be removed since
$\boldsymbol{x}_{i}^{k+1}$ is a random vector determined by
$\boldsymbol{a}_i^{k+1}$.

\subsubsection{Bounding the rest terms}
Regarding these terms, we have
\begin{align}
&\mathbb{E}_{\boldsymbol{a}_i^{k+1}}\big[(\boldsymbol{a}_i^{k+1}\odot
(\boldsymbol{x}_{i}^*-\boldsymbol{x}_{i}^{k+1}))^T(\boldsymbol{\lambda}_{i}^k+
\sigma_1(\boldsymbol{x}_{i}^{k+1}-\boldsymbol{H}_{i}^T
\boldsymbol{y}_{i}^k))
\nonumber\\
&\quad\qquad\big |\{\boldsymbol{a}_i^{t}\}\big]
\nonumber\\
&=\textstyle\mathbb{E}_{\boldsymbol{a}_i^{k+1}}\Big[\sum\limits_{j=1}^{|S_i|}
\hat{a}_{ij}^{k+1}\Big((\boldsymbol{x}_{ij}^*-\boldsymbol{x}_{ij}^{k})^T
\boldsymbol{\lambda}_{ij}^k+(\boldsymbol{x}_{ij}^{k}
-\boldsymbol{x}_{ij}^{k+1})^T\boldsymbol{\lambda}_{ij}^k
\nonumber\\
&\qquad\qquad+\sigma_1(\boldsymbol{x}_{ij}^*-\boldsymbol{x}_{ij}^{k+1})^T
(\boldsymbol{x}_{ij}^{k+1}-\boldsymbol{H}_{ij}^T\boldsymbol{y}_{i}^k)
\Big)\Big |\{\boldsymbol{a}_i^{t}\}\Big]
\nonumber\\
&\overset{(a)}{=}\textstyle\alpha(\boldsymbol{x}_{i}^{*}-\boldsymbol{x}_{i}^{k})^T
\boldsymbol{\lambda}_{i}^k+\mathbb{E}_{\boldsymbol{a}_i^{k+1}}
\Big[\sum\limits_{j=1}^{|S_i|}\hat{a}_{ij}^{k+1}\Big((\boldsymbol{x}_{ij}^{k}
-\boldsymbol{x}_{ij}^{k+1})^T\boldsymbol{\lambda}_{ij}^k
\nonumber\\
&\qquad\qquad+\sigma_1(\boldsymbol{x}_{ij}^*-\boldsymbol{x}_{ij}^{k+1})^T
(\boldsymbol{x}_{ij}^{k+1}-\boldsymbol{H}_{ij}^T\boldsymbol{y}_{i}^k)\Big)\Big |
\{\boldsymbol{a}_i^{t}\}\Big]
\nonumber\\
&\overset{(b)}{=}\textstyle(\alpha-1)(\boldsymbol{x}_{i}^{*}-\boldsymbol{x}_{i}^{k})^T
\boldsymbol{\lambda}_{i}^k+\mathbb{E}_{\boldsymbol{a}_i^{k+1}}\Big[(\boldsymbol{x}_{i}^{*}-
\boldsymbol{x}_{i}^{k+1})^T\boldsymbol{\lambda}_{i}^k+
\nonumber\\
&\textstyle\sum_{j=1}^{|S_i|}\hat{a}_{ij}^{k+1}
\sigma_1\Big((\boldsymbol{x}_{ij}^{*}-\boldsymbol{x}_{ij}^{k})^T
(\boldsymbol{x}_{ij}^{k}-\boldsymbol{H}_{ij}^T\boldsymbol{y}_{i}^k)+
(\boldsymbol{x}_{ij}^{k}-\boldsymbol{x}_{ij}^{k+1})^T
\nonumber\\
&(\boldsymbol{x}_{ij}^{k}-\boldsymbol{H}_{ij}^T\boldsymbol{y}_{i}^k)+
(\boldsymbol{x}_{ij}^{*}-\boldsymbol{x}_{ij}^{k+1})^T
(\boldsymbol{x}_{ij}^{k+1}-\boldsymbol{x}_{ij}^{k})\Big)
\Big|\{\boldsymbol{a}_i^{t}\}\Big]
\nonumber\\
&\overset{(c)}{=}(\alpha-1)(\boldsymbol{x}_{i}^{*}-\boldsymbol{x}_{i}^{k})^T
\boldsymbol{\lambda}_{i}^k+\underbrace{\mathbb{E}_{\boldsymbol{a}_i^{k+1}}
\big[(\boldsymbol{x}_{i}^{*}-\boldsymbol{x}_{i}^{k+1})^T\boldsymbol{\lambda}_{i}^k
\big|\{\boldsymbol{a}_i^{t}\}\big]}_{\text{[(\ref{appendix-2-3})-1]}}
\nonumber\\
&\quad+(\alpha-1)\sigma_1(\boldsymbol{x}_{i}^{*}-\boldsymbol{x}_{i}^{k})^T
(\boldsymbol{x}_{i}^{k}-\boldsymbol{H}_i^T\boldsymbol{y}_{i}^k)
\nonumber\\
&\quad+\underbrace{\textstyle\sigma_1\mathbb{E}_{\boldsymbol{a}_i^{k+1}}
\Big[(\boldsymbol{x}_{i}^{*}-\boldsymbol{x}_{i}^{k+1})^T
(\boldsymbol{x}_{i}^{k}-\boldsymbol{H}_i^T\boldsymbol{y}_{i}^k)
}_{\text{[(\ref{appendix-2-3})-2]}}
\nonumber\\
&\quad\underbrace{\textstyle+\sum_{j=1}^{|S_i|}\hat{a}_{ij}^{k+1}
(\boldsymbol{x}_{ij}^{*}-\boldsymbol{x}_{ij}^{k+1})^T
(\boldsymbol{x}_{ij}^{k+1}-\boldsymbol{x}_{ij}^{k})
\Big|\{\boldsymbol{a}_i^{t}\}\Big]}_{\text{[(\ref{appendix-2-3})-2]}}
\label{appendix-2-3}
\end{align}
where $(a)$ is because $\mathbb{E}_{\boldsymbol{a}_i^{k+1}}
\big[\sum_{j=1}^{|S_i|}\hat{a}_{ij}^{k+1}(\boldsymbol{x}_{ij}^*-
\boldsymbol{x}_{ij}^{k})^T\boldsymbol{\lambda}_{ij}^k\big|\{\boldsymbol{a}_i^{t}\}
\big]=\alpha(\boldsymbol{x}_{i}^{*}-\boldsymbol{x}_{i}^{k})^T
\boldsymbol{\lambda}_{i}^k$, $(b)$ and $(c)$ have invoked the same
logic as in (\ref{appendix-2-2-1}). Regarding $\text{[(\ref{appendix-2-3})-1]}$
and $\text{[(\ref{appendix-2-3})-2]}$, we have
\begin{align}
&\text{[(\ref{appendix-2-3})-1]}+\text{[(\ref{appendix-2-3})-2]}
\overset{(a)}{=}\mathbb{E}_{\boldsymbol{a}_i^{k+1}}
\Big[(\boldsymbol{x}_{i}^*-\boldsymbol{x}_{i}^{k+1})^T
\big(\bar{\boldsymbol{\lambda}}_{i}^{k+1}+
\nonumber\\
&\sigma_1(\boldsymbol{x}_{i}^{k}-\boldsymbol{x}_{i}^{k+1})+
\sigma_1\boldsymbol{H}_i^T(\boldsymbol{y}_{i}^{k+1}
-\boldsymbol{y}_{i}^{k})\big)+
\nonumber\\
&\textstyle\sigma_1\sum_{j=1}^{|S_i|}\hat{a}_{ij}^{k+1}
(\boldsymbol{x}_{ij}^{*}-\boldsymbol{x}_{ij}^{k+1})^T
(\boldsymbol{x}_{ij}^{k+1}-\boldsymbol{x}_{ij}^{k})
\Big|\{\boldsymbol{a}_i^{t}\}\Big]
\nonumber\\
&\overset{(b)}{=}\mathbb{E}_{\boldsymbol{a}_i^{k+1}}\big[(\boldsymbol{x}_{i}^*-
\boldsymbol{x}_{i}^{k+1})^T\big(\bar{\boldsymbol{\lambda}}_{i}^{k+1}+
\sigma_1\boldsymbol{H}_i^T(\boldsymbol{y}_{i}^{k+1}
-\boldsymbol{y}_{i}^{k})\big)\big|\{\boldsymbol{a}_i^{t}\}\big]
\nonumber\\
&\overset{(c)}{=}\mathbb{E}_{\boldsymbol{a}_i^{k+1}}
\Big[(\boldsymbol{x}_{i}^*-\boldsymbol{x}_{i}^{k+1})^T
\big(\alpha^{-1}(\boldsymbol{\lambda}_{i}^{k+1}-\boldsymbol{\lambda}_{i}^k)+
\boldsymbol{\lambda}_{i}^k
\nonumber\\
&\qquad\qquad+\sigma_1\boldsymbol{H}_i^T(\boldsymbol{y}_{i}^{k+1}
-\boldsymbol{y}_{i}^{k})\big)\big|\{\boldsymbol{a}_i^{t}\}\Big]
\nonumber\\
&=\mathbb{E}_{\boldsymbol{a}_i^{k+1}}\Big[(\boldsymbol{x}_{i}^*-
\boldsymbol{x}_{i}^{k+1})^T
\big(\boldsymbol{\lambda}_{i}^{k+1}+(1-\alpha^{-1})
(\boldsymbol{\lambda}_{i}^{k}-\boldsymbol{\lambda}_{i}^{k+1})
\nonumber\\
&\qquad\qquad+\sigma_1\boldsymbol{H}_i^T(\boldsymbol{y}_{i}^{k+1}-
\boldsymbol{y}_{i}^{k})\big)\big|\{\boldsymbol{a}_i^{t}\}\Big]
\nonumber\\
&\overset{(d)}{=}\mathbb{E}_{\boldsymbol{a}_i^{k+1}}\Big[(\boldsymbol{x}_{i}^*
-\boldsymbol{x}_{i}^{k+1})^T\boldsymbol{\lambda}_{i}^{k+1}+
\nonumber\\
&\qquad\qquad(1-\alpha)\sigma_1(\boldsymbol{x}_{i}^*-\boldsymbol{x}_{i}^{k+1})^T
(\boldsymbol{x}_{i}^{k+1}-\boldsymbol{H}_i^T\boldsymbol{y}_{i}^{k+1})
\nonumber\\
&\qquad\qquad+\sigma_1(\boldsymbol{x}_{i}^*-\boldsymbol{x}_{i}^{k+1})^T
\boldsymbol{H}_i^T(\boldsymbol{y}_{i}^{k+1}-\boldsymbol{y}_{i}^{k})
\big|\{\boldsymbol{a}_i^{t}\}\Big],
\label{appendix-2-5}
\end{align}
where $\bar{\boldsymbol{\lambda}}_{i}^{k+1}$ is defined in (\ref{sec2-2}),
$(a)$ and $(c)$ have invoked (\ref{sec2-2}), $(b)$ is because
\begin{align}
&\textstyle(\boldsymbol{x}_{i}^*-\boldsymbol{x}_{i}^{k+1})^T(\boldsymbol{x}_{i}^{k}-
\boldsymbol{x}_{i}^{k+1})+
\nonumber\\
&\textstyle\sum_{j=1}^{|S_i|}\hat{a}_{ij}^{k+1}(\boldsymbol{x}_{ij}^{*}-
\boldsymbol{x}_{ij}^{k+1})^T(\boldsymbol{x}_{ij}^{k+1}-\boldsymbol{x}_{ij}^{k})=0,
\end{align}
since $\boldsymbol{x}_{ij}^{k+1}=\boldsymbol{x}_{ij}^{k}$,
$\forall j \notin I_i^{k+1}$, and $(d)$ is due to
\begin{align}
\boldsymbol{\lambda}_{i}^{k}-\boldsymbol{\lambda}_{i}^{k+1}
\overset{(e)}{=}-\alpha(\bar{\boldsymbol{\lambda}}_{i}^{k+1}
-\boldsymbol{\lambda}_{i}^k)
\overset{(f)}{=}-\alpha\sigma_1(\boldsymbol{x}_{i}^{k+1}-
\boldsymbol{H}_i^T\boldsymbol{y}_{i}^{k+1}). \nonumber
\end{align}
in which $(e)$ and $(f)$ come from the second line and the first
line of (\ref{sec3-2-1-1}), respectively. Substituting
(\ref{appendix-2-2}), (\ref{appendix-2-3}) and
(\ref{appendix-2-5}) into (\ref{appendix-2-1}) yields
\begin{align}
&0\leq(\alpha-1)(F_i^k+M_i^k+G_i^k)+\mathbb{E}_{\boldsymbol{a}_i^{k+1}}
\Big[F_i^{k+1}+M_i^{k+1}+
\nonumber\\
&(1-\alpha)G_i^{k+1}+T_i^{k+1}-
\nonumber\\
&\underbrace{\textstyle\sum\limits_{j=1}^{|S_i|}\hat{a}_{ij}^{k+1}
\big((\boldsymbol{x}_{ij}^*-\boldsymbol{x}_{ij}^{k+1})^T\boldsymbol{\tau}_{ij}^{k+1}
+\frac{\mu}{2}\big\|\boldsymbol{x}_{ij}^{*}-
\boldsymbol{x}_{ij}^{k+1}\big\|_2^2\big)}_{\text{[(\ref{appendix-2-6})-1]}}
\Big|\{\boldsymbol{a}_i^{t}\}\Big],
\label{appendix-2-6}
\end{align}
where $F_i^k\triangleq f_i(\boldsymbol{x}_i^{*})-f_i(\boldsymbol{x}_i^{k})$,
$M_i^k\triangleq(\boldsymbol{x}_{i}^{*}-\boldsymbol{x}_{i}^{k})^T
\boldsymbol{\lambda}_{i}^k$,
\begin{align}
&G_i^k\triangleq\sigma_1(\boldsymbol{x}_{i}^{*}-\boldsymbol{x}_{i}^{k})^T
(\boldsymbol{x}_{i}^{k}-\boldsymbol{H}_{i}^T\boldsymbol{y}_{i}^k),
\nonumber\\
&T_i^{k+1}\triangleq\sigma_1(\boldsymbol{x}_{i}^{*}-\boldsymbol{x}_{i}^{k+1})^T
\boldsymbol{H}_i^T(\boldsymbol{y}_{i}^{k+1}-\boldsymbol{y}_{i}^{k}),
\label{appendix-2-6-1}
\end{align}
and the left hand side of (\ref{appendix-2-1}) has been moved to the right
hand side of (\ref{appendix-2-6}), i.e., the first term in
$\text{[(\ref{appendix-2-6})-1]}$. Regarding $\text{[(\ref{appendix-2-6})-1]}$,
we have
\begin{align}
&\text{[(\ref{appendix-2-6})-1]}\textstyle\overset{(a)}{\leq}
\sum_{j=1}^{|S_i|}\hat{a}_{ij}^{k+1}\big(\frac{\mu}{2}
\|\boldsymbol{x}_{ij}^*-\boldsymbol{x}_{ij}^{k+1}\|_2^2+\frac{1}{2\mu}
\|\boldsymbol{\tau}_{ij}^{k+1}\|_2^2-
\nonumber\\
&\textstyle\frac{\mu}{2}\big\|\boldsymbol{x}_{ij}^*-\boldsymbol{x}_{ij}^{k+1}
\big\|_2^2\big)\overset{(b)}{\leq}\textstyle \frac{(\epsilon^{k+1})^2
\sum_{j=1}^{|S_i|}\hat{a}_{ij}^{k+1}}{2\mu},
\label{appendix-2-7}
\end{align}
where $(a)$ has invoked (\ref{sec2-2-2-1}) and $(b)$ is because
$\|\boldsymbol{\tau}_{ij}^{k+1}\|_2\leq \epsilon^{k+1}$,
see (\ref{imple-1}). Substituting (\ref{appendix-2-7}) into (\ref{appendix-2-6})
and also using the fact that $\mathbb{E}_{\boldsymbol{a}_i^{k+1}}
[\sum_{j=1}^{|S_i|}\hat{a}_{ij}^{k+1}]=\alpha|S_i|$
yields the desired result.

\section{Proving $R\leq \textstyle\tilde{C}^0\big(\{\boldsymbol{\lambda}_i\},\boldsymbol{\beta}
\big)$}
\label{sec-appendix-3}
First notice that
\begin{align}
&\textstyle R=\sum_{i=1}^l \Big( C_i^0+\mathbb{E}
\Big[\sum_{t=1}^{\bar{k}}T_i^{t}+\textstyle\frac{1}{\sigma_1}
(\bar{\boldsymbol{\lambda}}_i^{\bar{k}}-\boldsymbol{\lambda}_i)^T
(\boldsymbol{\lambda}_{i}^{\bar{k}-1}-\bar{\boldsymbol{\lambda}}_{i}^{\bar{k}})
\nonumber\\
&\textstyle+\frac{1}{\sigma_1}
\sum_{t=1}^{\bar{k}-1}(\boldsymbol{\lambda}_i^{t}-
\boldsymbol{\lambda}_i)^T(\boldsymbol{\lambda}_{i}^{t-1}-
\boldsymbol{\lambda}_{i}^{t})\Big]\Big)+\sigma_2\sum_{t=1}^{\bar{k}}
F^{t}_{(\boldsymbol{P},\boldsymbol{y})}+
\nonumber\\
&\textstyle\frac{\alpha}{\sigma_2}\sum_{t=1}^{\bar{k}}
(\boldsymbol{\beta}-\boldsymbol{\beta}^{t})^T(\boldsymbol{\beta}^{t}-
\boldsymbol{\beta}^{t-1}).
\end{align}
We then separately upper bound some of the terms in $R$ to prove the claim.

\subsubsection{Bounding $T_i^{\bar{k}}$}
Regarding this term, first notice that
\begin{align}
&\bar{\boldsymbol{\lambda}}_i^{\bar{k}}\overset{(a)}{=}
\boldsymbol{\lambda}_i^{\bar{k}-1}+\sigma_1(\boldsymbol{x}_i^{\bar{k}}-
\boldsymbol{H}_i^T\boldsymbol{y}_i^{\bar{k}}) \
\overset{\boldsymbol{x}_i^*-\boldsymbol{H}_i^T\boldsymbol{y}_i^*
=\boldsymbol{0}}{\Rightarrow}
\nonumber\\
&\boldsymbol{x}_i^*-\boldsymbol{x}_i^{\bar{k}}=
-\sigma_1^{-1}(\bar{\boldsymbol{\lambda}}_i^{\bar{k}}-
\boldsymbol{\lambda}_i^{\bar{k}-1})
+\boldsymbol{H}_i^T(\boldsymbol{y}_i^*-\boldsymbol{y}_i^{\bar{k}}).
\label{appendix-3-1}
\end{align}
where $(a)$ comes from (\ref{sec2-2}). Thus we have
\begin{align}
&T_i^{\bar{k}}
=-(\bar{\boldsymbol{\lambda}}_i^{\bar{k}}-
\boldsymbol{\lambda}_i^{\bar{k}-1})^T\boldsymbol{H}_i^T
(\boldsymbol{y}_{i}^{\bar{k}}-\boldsymbol{y}_{i}^{\bar{k}-1})
\nonumber\\
&\qquad+\sigma_1(\boldsymbol{H}_i^T(\boldsymbol{y}_{i}^{*}-
\boldsymbol{y}_{i}^{\bar{k}}))^T\boldsymbol{H}_i^T
(\boldsymbol{y}_{i}^{\bar{k}}-\boldsymbol{y}_{i}^{\bar{k}-1})
\nonumber\\
&\overset{(a)}{=}\textstyle\underbrace{-(\bar{\boldsymbol{\lambda}}_i^{\bar{k}}
-\boldsymbol{\lambda}_i^{\bar{k}-1})^T\boldsymbol{H}_i^T
(\boldsymbol{y}_{i}^{\bar{k}}-\boldsymbol{y}_{i}^{\bar{k}-1})}_{
\text{[(\ref{appendix-3-2})-1]}}-\Big(\frac{\sigma_1}{2}
\|\boldsymbol{H}_i^T(\boldsymbol{y}_{i}^{*}-\boldsymbol{y}_{i}^{\bar{k}})\|_2^2
\nonumber\\
&\qquad\underbrace{+\textstyle\frac{\sigma_1}{2}\|\boldsymbol{H}_i^T
(\boldsymbol{y}_{i}^{\bar{k}}-\boldsymbol{y}_{i}^{\bar{k}-1})\|_2^2}_{
\text{[(\ref{appendix-3-2})-2]}}-\textstyle\frac{\sigma_1}{2}\|\boldsymbol{H}_i^T
(\boldsymbol{y}_{i}^{*}-\boldsymbol{y}_{i}^{\bar{k}-1})\|_2^2\Big)
\nonumber\\
&\overset{(b)}{\leq}\textstyle \underbrace{\textstyle\frac{1}{2\sigma_1}
\|\bar{\boldsymbol{\lambda}}_i^{\bar{k}}-\boldsymbol{\lambda}_i^{\bar{k}-1}
\|_2^2}_{\text{[(\ref{appendix-3-2})-3]}}-
\nonumber\\
&\qquad\underbrace{\textstyle
\frac{\sigma_1}{2}(\|\boldsymbol{H}_i^T(\boldsymbol{y}_{i}^{*}-
\boldsymbol{y}_{i}^{\bar{k}})\|_2^2-\|\boldsymbol{H}_i^T(\boldsymbol{y}_{i}^{*}-
\boldsymbol{y}_{i}^{\bar{k}-1})\|_2^2)}_{\text{[(\ref{appendix-3-2})-4]}}
\label{appendix-3-2}
\end{align}
where $(a)$ has invoked (\ref{sec2-2-2}) and $(b)$ is because
$\text{[(\ref{appendix-3-2})-1]}-\text{[(\ref{appendix-3-2})-2]}-
\text{[(\ref{appendix-3-2})-3]}\leq 0$.

\subsubsection{Bounding $T_i^t$, $t<\bar{k}$}
Similar to (\ref{appendix-3-1}), we can deduce that
\begin{align}
\textstyle\boldsymbol{x}_{i}^{*}-\boldsymbol{x}_{i}^{t}=
-\frac{1}{\alpha\sigma_1}(\boldsymbol{\lambda}_i^{t}-
\boldsymbol{\lambda}_i^{t-1})+\boldsymbol{H}_i^T
(\boldsymbol{y}_{i}^{*}-\boldsymbol{y}_{i}^{t}), \ t<\bar{k},
\label{appendix-3-2-1}
\end{align}
where the inequality is due to (\ref{sec2-2}) as well as the
fact that $\boldsymbol{x}_i^*-\boldsymbol{H}_i^T\boldsymbol{y}_i^*
=\boldsymbol{0}$. Substituting the right hand side of
(\ref{appendix-3-2-1}) into $T_i^t$, we have
\begin{align}
&T_i^t\textstyle=-\frac{1}{\alpha}(\boldsymbol{\lambda}_i^{t}
-\boldsymbol{\lambda}_i^{t-1})^T\boldsymbol{H}_i^T(\boldsymbol{y}_{i}^{t}-
\boldsymbol{y}_{i}^{t-1})+
\nonumber\\
&\qquad\sigma_1(\boldsymbol{H}_i^T(\boldsymbol{y}_{i}^{*}-
\boldsymbol{y}_{i}^{t}))^T\boldsymbol{H}_i^T(\boldsymbol{y}_{i}^{t}-
\boldsymbol{y}_{i}^{t-1})
\nonumber\\
&\textstyle\overset{(a)}{=}-\frac{1}{\alpha}(\boldsymbol{\lambda}_i^{t}
-\boldsymbol{\lambda}_i^{t-1})^T\boldsymbol{H}_i^T(\boldsymbol{y}_{i}^{t}-
\boldsymbol{y}_{i}^{t-1})-\frac{\sigma_1}{2}\big(\|\boldsymbol{H}_i^T
(\boldsymbol{y}_{i}^{*}-\boldsymbol{y}_{i}^{t})\|_2^2
\nonumber\\
&\quad+\|\boldsymbol{H}_i^T(\boldsymbol{y}_{i}^{t}-\boldsymbol{y}_{i}^{t-1})\|_2^2
-\|\boldsymbol{H}_i^T(\boldsymbol{y}_{i}^{*}-\boldsymbol{y}_{i}^{t-1})\|_2^2\big)
\nonumber\\
&\overset{(b)}{\leq} \underbrace{\textstyle\frac{1}{2\sigma_1}
\|\boldsymbol{\lambda}_i^{t}-\boldsymbol{\lambda}_i^{t-1}\|_2^2+
\left(\frac{\sigma_1}{2\alpha^2}-\frac{\sigma_1}{2}\right)\|\boldsymbol{H}_i^T
(\boldsymbol{y}_{i}^{t}-\boldsymbol{y}_{i}^{t-1})\|_2^2-}_{\text{[(\ref{appendix-3-3})-1]}}
\nonumber\\
&\underbrace{\textstyle\frac{\sigma_1}{2}(\|\boldsymbol{H}_i^T(\boldsymbol{y}_{i}^{*}-
\boldsymbol{y}_{i}^{t})\|_2^2-\|\boldsymbol{H}_i^T(\boldsymbol{y}_{i}^{*}-
\boldsymbol{y}_{i}^{t-1})\|_2^2)}_{\text{[(\ref{appendix-3-3})-1]}}
\label{appendix-3-3}
\end{align}
where $(a)$ and $(b)$ are obtained similarly as in (\ref{appendix-3-2}).

\subsubsection{Bounding the rest terms}
Regarding these terms, by (\ref{sec2-2-2}) we have
\begin{align}
&(\bar{\boldsymbol{\lambda}}_i^{\bar{k}}-\boldsymbol{\lambda}_i)^T
(\boldsymbol{\lambda}_{i}^{\bar{k}-1}-\bar{\boldsymbol{\lambda}}_{i}^{\bar{k}})=
\nonumber\\
&\underbrace{-0.5(\|\bar{\boldsymbol{\lambda}}_i^{\bar{k}}-
\boldsymbol{\lambda}_i\|_2^2+\|\boldsymbol{\lambda}_{i}^{\bar{k}-1}-
\bar{\boldsymbol{\lambda}}_{i}^{\bar{k}}\|_2^2-
\|\boldsymbol{\lambda}_i^{\bar{k}-1}-\boldsymbol{\lambda}_i
\|_2^2)}_{\text{[(\ref{appendix-3-4})-1]}},
\nonumber\\
&(\boldsymbol{\lambda}_i^{t}-\boldsymbol{\lambda}_i)^T
(\boldsymbol{\lambda}_{i}^{t-1}-\boldsymbol{\lambda}_{i}^{t})=
\nonumber\\
&\underbrace{-0.5(\|\boldsymbol{\lambda}_i^{t}-\boldsymbol{\lambda}_i\|_2^2+
\|\boldsymbol{\lambda}_{i}^{t-1}-\boldsymbol{\lambda}_{i}^{t}\|_2^2-
\|\boldsymbol{\lambda}_i^{t-1}-\boldsymbol{\lambda}_i\|_2^2
)}_{\text{[(\ref{appendix-3-4})-2]}},
t<\bar{k},
\nonumber\\
&\textstyle F_{\boldsymbol{P},\boldsymbol{y}}^t=
\nonumber\\
&\underbrace{-0.5(\|\boldsymbol{y}^*-
\boldsymbol{y}^{t}\|_{\boldsymbol{P}}^2+\|\boldsymbol{y}^{t}-
\boldsymbol{y}^{t-1}\|_{\boldsymbol{P}}^2-\|\boldsymbol{y}^*-
\boldsymbol{y}^{t-1}\|_{\boldsymbol{P}}^2
)}_{\text{[(\ref{appendix-3-4})-3]}},\forall t,
\nonumber\\
&\textstyle(\boldsymbol{\beta}-\boldsymbol{\beta}^{t})^T
(\boldsymbol{\beta}^{t}-\boldsymbol{\beta}^{t-1})=
\nonumber\\
&\underbrace{-0.5(\|\boldsymbol{\beta}-
\boldsymbol{\beta}^{t}\|_2^2+\|\boldsymbol{\beta}^{t}-
\boldsymbol{\beta}^{t-1}\|_2^2-\|\boldsymbol{\beta}-
\boldsymbol{\beta}^{t-1}\|_2^2)}_{\text{[(\ref{appendix-3-4})-4]}},
\forall t.
\label{appendix-3-4}
\end{align}

\subsubsection{Combining}
Substituting (\ref{appendix-3-2}), (\ref{appendix-3-3}) and (\ref{appendix-3-4})
into $R$, we have
\begin{align}
&\textstyle R\leq \sum_{i=1}^lC_i^0+\mathbb{E}\big[\sum_{i=1}^l
\text{[(\ref{appendix-3-2})-3]}-\text{[(\ref{appendix-3-2})-4]}\big]+
\nonumber\\
&\textstyle\mathbb{E}\big[\sum_{i=1}^l\sum_{t=1}^{\bar{k}-1}
\text{[(\ref{appendix-3-3})-1]}\big]+\frac{1}{\sigma_1}\mathbb{E}
\big[\sum_{i=1}^l\text{[(\ref{appendix-3-4})-1]}\big]+
\nonumber\\
&\textstyle\frac{1}{\sigma_1}\mathbb{E}\big[\sum_{i=1}^l\sum_{t=1}^{\bar{k}-1}
\text{[(\ref{appendix-3-4})-2]}\big]+\mathbb{E}\big[\sum_{t=1}^{\bar{k}}
\sigma_2\text{[(\ref{appendix-3-4})-3]}+\frac{\alpha}{\sigma_2}
\text{[(\ref{appendix-3-4})-4]}\big].
\label{appendix-3-7}
\end{align}
Eliminating the repeated terms in the right hand side of (\ref{appendix-3-7})
leads to
\begin{align}
&R\leq\textstyle\sum_{i=1}^lC_i^0+
\textstyle\mathbb{E}\Big[\sum_{i=1}^l
\Big(\frac{\sigma_1}{2}\big(\|\boldsymbol{H}_i^T
(\boldsymbol{y}_{i}^{*}-\boldsymbol{y}_{i}^{0})\|_2^2-
\nonumber\\
&\textstyle\|\boldsymbol{H}_i^T(\boldsymbol{y}_{i}^{*}-
\boldsymbol{y}_{i}^{\bar{k}})\|_2^2\big)+\frac{1}{2\sigma_1}
\big(\|\boldsymbol{\lambda}_i^{0}-\boldsymbol{\lambda}_i\|_2^2-
\|\bar{\boldsymbol{\lambda}}_i^{\bar{k}}-\boldsymbol{\lambda}_i\|_2^2\big)\Big)
\nonumber\\
&\textstyle+(\frac{\sigma_1}{2\alpha^2}-\frac{\sigma_1}{2})
\sum_{i=1}^l\sum_{t=1}^{\bar{k}-1}\|\boldsymbol{H}_i^T(\boldsymbol{y}_{i}^{t}
-\boldsymbol{y}_{i}^{t-1})\|_2^2-
\nonumber\\
&\textstyle\frac{\sigma_2}{2}\sum_{t=1}^{\bar{k}}\|\boldsymbol{y}^{t}-
\boldsymbol{y}^{t-1}\|_{\boldsymbol{P}}^2
-\frac{\sigma_2}{2}(\|\boldsymbol{y}^{*}-\boldsymbol{y}^{\bar{k}}
\|_{\boldsymbol{P}}^2-\|\boldsymbol{y}^{*}-\boldsymbol{y}^{0}
\|_{\boldsymbol{P}}^2)-
\nonumber\\
&\textstyle\frac{\alpha}{2\sigma_2}
\sum_{t=1}^{\bar{k}}\|\boldsymbol{\beta}^{t}-\boldsymbol{\beta}^{t-1}\|_2^2
-\frac{\alpha}{2\sigma_2}(\|\boldsymbol{\beta}-\boldsymbol{\beta}^{\bar{k}}\|_2^2
-\|\boldsymbol{\beta}-\boldsymbol{\beta}^{0}\|_2^2)\Big]
\nonumber\\
&\leq C^0\big(\{\boldsymbol{\lambda}_i\},\boldsymbol{\beta}\big)+
C^2,
\label{appendix-3-8}
\end{align}
where the second inequality is obtained by defining
\begin{align}
&\textstyle C^0\big(\{\boldsymbol{\lambda}_i\},\boldsymbol{\beta}\big)
\triangleq \sum_{i=1}^lC_i^0+\Big(\sum_{i=1}^l\frac{\sigma_1}{2}
\|\boldsymbol{H}_i^T(\boldsymbol{y}_{i}^{*}-\boldsymbol{y}_{i}^{0})\|_2^2
\nonumber\\
&\textstyle+\frac{1}{2\sigma_1}\|\boldsymbol{\lambda}_i^{0}-\boldsymbol{\lambda}_i\|_2^2\Big)
+\frac{\sigma_2}{2}\|\boldsymbol{y}^{*}-\boldsymbol{y}^{0}\|_{\boldsymbol{P}}^2
+\frac{\alpha}{2\sigma_2}\|\boldsymbol{\beta}-\boldsymbol{\beta}^{0}\|_2^2,
\nonumber\\
&C^2\triangleq\textstyle\mathbb{E}\Big[\sum_{i=1}^l\sum_{t=1}^{\bar{k}-1}
\left(\frac{\sigma_1}{2\alpha^2}-\frac{\sigma_1}{2}\right)
\|\boldsymbol{H}_i^T(\boldsymbol{y}_{i}^{t}-\boldsymbol{y}_{i}^{t-1})\|_2^2
\nonumber\\
&\textstyle-\sum_{t=1}^{\bar{k}}\big(\frac{\sigma_2}{2}\|\boldsymbol{y}^{t}-
\boldsymbol{y}^{t-1}\|_{\boldsymbol{P}}^2+\frac{\alpha}{2\sigma_2}
\|\boldsymbol{\beta}^{t}-\boldsymbol{\beta}^{t-1}\|_2^2\big)\Big]
\end{align}
and also by omitting some negative terms in the right hand side of the first
inequality. Next, we separately bound the terms in $C^2$. According to the
definitions of $\boldsymbol{H}_i$ and $\boldsymbol{H}$, i.e. (\ref{sec3-2-1}),
it holds
\begin{align}
&\textstyle\sum_{i=1}^l\left(\frac{\sigma_1}{2\alpha^2}-\frac{\sigma_1}{2}\right)
\|\boldsymbol{H}_i^T(\boldsymbol{y}_{i}^{t}-\boldsymbol{y}_{i}^{t-1})\|_2^2
\nonumber\\
=&\textstyle\left(\frac{\sigma_1}{2\alpha^2}-\frac{\sigma_1}{2}\right)
\|\boldsymbol{H}^T(\boldsymbol{y}^{t}-\boldsymbol{y}^{t-1})\|_2^2.
\label{appendix-5-1}
\end{align}
Meanwhile, regarding $\sum_{t=1}^{\bar{k}-1}\|\boldsymbol{\beta}^{t}-
\boldsymbol{\beta}^{t-1}\|_2^2$ we have
\begin{align}
&\textstyle\sum_{t=1}^{\bar{k}-1}\|\boldsymbol{\beta}^{t}-
\boldsymbol{\beta}^{t-1}\|_2^2\overset{(a)}{=}
\sigma_2^2\sum_{t=1}^{\bar{k}-1}\|\boldsymbol{A}\boldsymbol{y}^{t}\|_2^2
\nonumber\\
\geq & \textstyle\frac{\sigma_2^2}{2}
\sum_{t=1}^{\bar{k}-1}\left(\|\boldsymbol{A}\boldsymbol{y}^{t}\|_2^2+
\|\boldsymbol{A}\boldsymbol{y}^{t-1}\|_2^2\right)-\frac{\sigma_2^2}{2}
\|\boldsymbol{A}\boldsymbol{y}^{0}\|_2^2
\nonumber\\
\overset{(b)}{\geq}&\textstyle\frac{\sigma_2^2}{4}\sum_{t=1}^{\bar{k}-1}
\|\boldsymbol{A}(\boldsymbol{y}^{t}-\boldsymbol{y}^{t-1})\|_2^2
-\frac{\sigma_2^2}{2}\|\boldsymbol{A}\boldsymbol{y}^{0}\|_2^2
\label{appendix-5-2}
\end{align}
where $(a)$ comes from (\ref{sec2-2-y}) and $(b)$ is
due to the fact that $\|\boldsymbol{x}+\boldsymbol{y}\|_2^2\leq 2
\|\boldsymbol{x}\|_2^2+2\|\boldsymbol{y}\|_2^2$,
$\forall \boldsymbol{x},\boldsymbol{y}$. Substituting
(\ref{appendix-5-1}) and (\ref{appendix-5-2}) into $C^2$ yields
\begin{align}
&\textstyle C^2 \leq -\big(\frac{\sigma_2}{2}\|\boldsymbol{y}^{\bar{k}}-
\boldsymbol{y}^{\bar{k}-1}\|_{\boldsymbol{P}}^2+\frac{\alpha}{2\sigma_2}
\|\boldsymbol{\beta}^{\bar{k}}-\boldsymbol{\beta}^{\bar{k}-1}\|_2^2\big)
\nonumber\\
&+\textstyle(\frac{\sigma_1}{2\alpha^2}-
\frac{\sigma_1}{2})\sum\limits_{t=1}^{\bar{k}-1}\|\boldsymbol{H}^T
(\boldsymbol{y}^{t}-\boldsymbol{y}^{t-1})\|_2^2-\frac{\sigma_2}{2}
\sum\limits_{t=1}^{\bar{k}-1}\|\boldsymbol{y}^{t}-
\boldsymbol{y}^{t-1}\|_{\boldsymbol{P}}^2
\nonumber\\
&\textstyle
-\frac{\alpha}{2}\sum_{t=1}^{\bar{k}-1}\frac{\sigma_2}{4}
\|\boldsymbol{A}(\boldsymbol{y}^{t}-\boldsymbol{y}^{t-1})\|_2^2
+\frac{\alpha\sigma_2}{4}\|\boldsymbol{A}\boldsymbol{y}^{0}\|_2^2
\nonumber\\
& \leq \textstyle\frac{\alpha\sigma_2}{4}\|\boldsymbol{A}\boldsymbol{y}^{0}\|_2^2
\label{appendix-5-3}
\end{align}
where the second inequality is due to the condition imposed on
$\boldsymbol{P}$, i.e., (\ref{the-main1-1}). Substituting
(\ref{appendix-5-3}) into (\ref{appendix-3-8}), and defining
$\tilde{C}^0\big(\{\boldsymbol{\lambda}_i\},
\boldsymbol{\beta}\big)\triangleq
C^0\big(\{\boldsymbol{\lambda}_i\},
\boldsymbol{\beta}\big)+\frac{\alpha\sigma_2}{4}\|\boldsymbol{A}\boldsymbol{y}^{0}
\|_2^2$, we obtain the desired result.

\bibliography{main1}
\bibliographystyle{IEEEtran}

\end{document}